\documentclass[10pt,twocolumn,letterpaper]{article}

\usepackage[pagenumbers]{cvpr} 

\definecolor{cvprblue}{rgb}{0.21,0.49,0.74}
\usepackage[pagebackref,breaklinks,colorlinks,allcolors=cvprblue]{hyperref}

\usepackage{hyperref}       
\usepackage{url}            
\usepackage{booktabs}       
\usepackage{amsfonts}       
\usepackage{nicefrac}       
\usepackage{microtype}      
\usepackage{xcolor}         
\definecolor{aliceblue}{RGB}{240,248,255}
\usepackage{colortbl}
\usepackage{amsmath} 
\usepackage {amssymb}

\usepackage{graphicx}
\usepackage[T1]{fontenc}

\usepackage{tikz,graphics,color,float,epsf,caption,subcaption}
\usepackage{booktabs, 
            makecell, multirow, tabularx}
\usepackage{algorithm}
\usepackage{algorithmic}

\def\paperID{3312} 
\def\confName{CVPR}
\def\confYear{2025}

\title{VARGPT-v1.1: Improve Visual Autoregressive Large Unified Model via Iterative Instruction Tuning and Reinforcement Learning }

\author{Xianwei Zhuang\textsuperscript{1}\thanks{Equal contribution}~,  Yuxin Xie~\textsuperscript{1}$^*$,  Yufan Deng\textsuperscript{1}$^*$, Dongchao Yang\textsuperscript{2}, \\Liming Liang\textsuperscript{1}, Jinghan Ru\textsuperscript{1}, Yuguo Yin\textsuperscript{1}, Yuexian Zou\textsuperscript{1}~\thanks{Corresponding author}\\
\textsuperscript{1}School of Electronic and Computer Engineering, Peking University\\
\textsuperscript{2}The Chinese University of Hong Kong\\
{\tt\small \{xwzhuang, yuxinxie, dengyufan10\}@stu.pku.edu.cn} \\
Technical Report \\
\textcolor{red}{Project Page: \url{https://vargpt1-1.github.io/}}\\
\textcolor{red}{Codes: \url{https://github.com/VARGPT-family/VARGPT-v1.1}}
}

\begin{document}
\maketitle

\begin{abstract}

In this work, we present VARGPT-v1.1, an advanced unified visual autoregressive model that builds upon our previous framework VARGPT. The model preserves the dual paradigm of next-token prediction for visual understanding and next-scale generation for image synthesis. Specifically, VARGPT-v1.1 integrates: (1) a novel training strategy combining iterative visual instruction tuning with reinforcement learning through Direct Preference Optimization (DPO), (2) an expanded training corpus containing 8.3M visual-generative instruction pairs, (3) an upgraded language model backbone using Qwen2, (4) enhanced image generation resolution, and (5) emergent image editing capabilities without architectural modifications.
These advancements enable VARGPT-v1.1 to achieve state-of-the-art performance in multimodal understanding and text-to-image instruction-following tasks, demonstrating significant improvements in both comprehension and generation metrics. Notably, through visual instruction tuning, the model acquires image editing functionality while maintaining architectural consistency with its predecessor, revealing the potential for unified visual understanding, generation, and editing. Our findings suggest that well-designed unified visual autoregressive models can effectively adopt flexible training strategies from large language models (LLMs), exhibiting promising scalability. The codebase and model weights are publicly available at \url{https://github.com/VARGPT-family/VARGPT-v1.1}.

\end{abstract}

\section{Introduction}
\label{sec:intro}

\begin{figure}[t]
   \centering
    \includegraphics[width=1\linewidth]{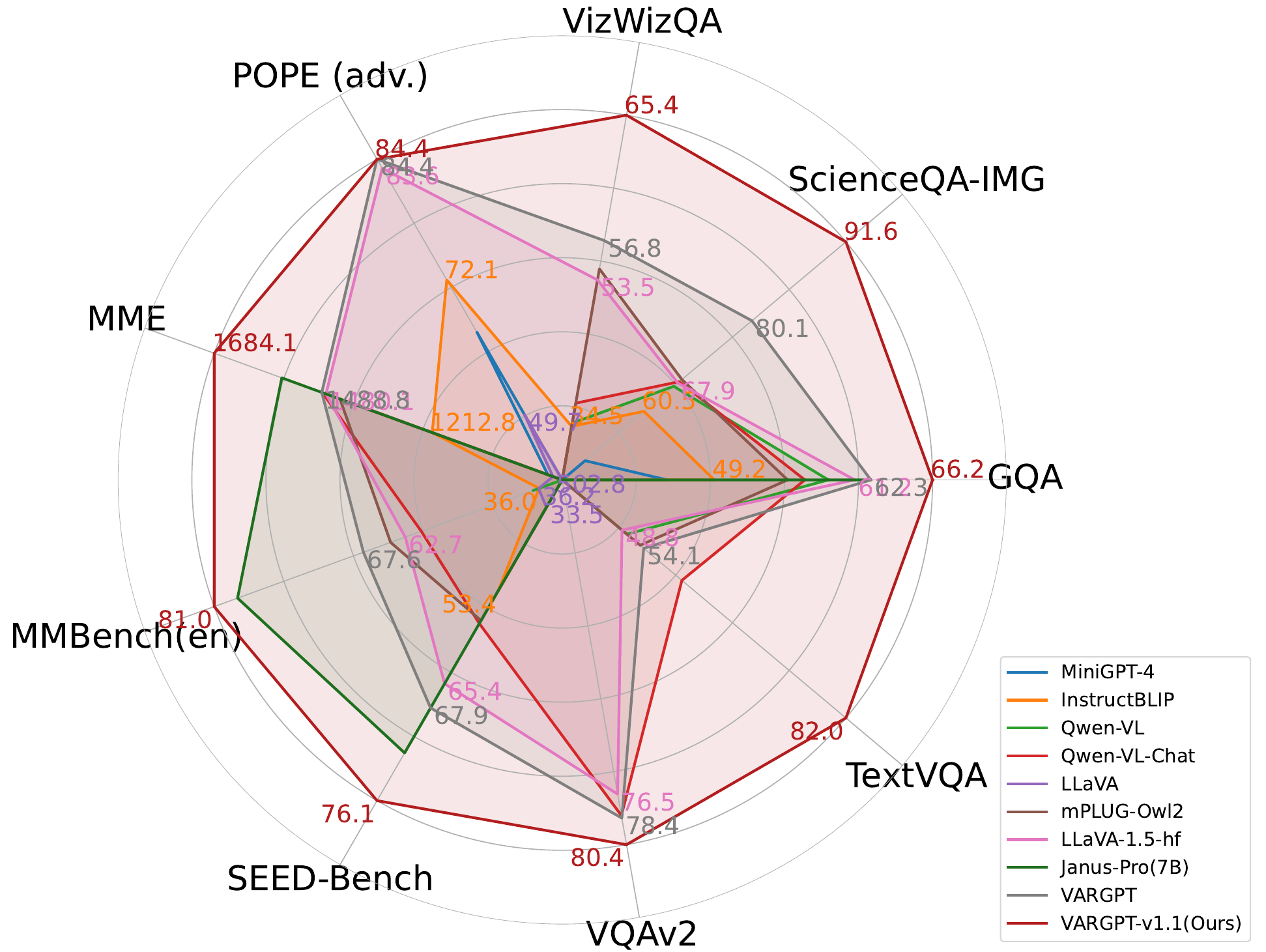}
    \caption{A comparative analysis of various MLLMs across multiple visual comprehension benchmarks is presented. The remaining metrics are derived from standard visual question-answering benchmarks and multi-modal comprehension benchmarks. Notably, our VARGPT-v1.1 model demonstrates significant superiority over the compared baselines across all comprehension benchmarks.}
    \label{fig:intro}
 \end{figure}

Recent advancements in multimodal AI have achieved dual breakthroughs in comprehension and generation. While Multimodal Large Language Models (MLLMs)~\cite{Liu2023VisualIT, Chen2023MiniGPTv2LL, Ye2023mPLUGOwl2RM, dai2023instructblipgeneralpurposevisionlanguagemodels, bai2023qwen} excel at cross-modal understanding, Denoising Diffusion Models~\cite{ho2020denoising,nichol2021improved} dominate visual generation through iterative refinement.
This progress drives the pursuit of unified frameworks (Figure~\ref{fig:vargpt_arch}), where three primary paradigms emerge: (1) \textit{Assembly Systems} integrating LLMs with diffusion models~\cite{wu2023next,ge2024seed,xie2024showo}, (2) \textit{Pure Autoregression} architectures predicting visual tokens~\cite{liu2023world,Chameleon_Team_Chameleon_Mixed-Modal_Early-Fusion_2024,wu2024janusdecouplingvisualencoding,chen2025janusprounifiedmultimodalunderstanding}, and (3) \textit{Dual-diffusion Models} with parallel generation mechanisms~\cite{li2024dualdiffusionunifiedimage}. Current implementations face representation conflicts between understanding and generation tasks, while TokenFlow~\cite{qu2024tokenflowunifiedimagetokenizer} unifies tokenization, its visual generation and understanding pipeline remains decoupled. Emerging solutions~\cite{wang2024emu3,VAR,sun2024autoregressive,tang2024hart,Infinity,wu2024liquid, wu2024janusdecouplingvisualencoding, chen2025janusprounifiedmultimodalunderstanding} address this through architectural innovations that preserve task-specific representations while enabling joint optimization

VARGPT~\cite{zhuang2025vargptunifiedunderstandinggeneration} endeavors to unify visual generation and understanding within a visual autoregressive MLLM, naturally supporting mixed-modal input and output. VARGPT models understanding and generation as two distinct paradigms within a unified model: {predicting the next token for visual understanding and predicting the next scale for visual generation}, respectively. 
However, due to the limited scale of training data and relatively insufficient training iterations, such approaches exhibit certain limitations, including suboptimal performance in instruction-following capabilities for image generation and constrained coverage of image generation domains (primarily limited to ImageNet~\cite{deng2009imagenet} categories).
To address these limitations, we present VARGPT-v1.1, which implements comprehensive enhancements over its predecessor while preserving the fundamental paradigm of VARGPT. Specifically, VARGPT-v1.1 achieves across-the-board improvements in both generative and comprehension capabilities through several  key technical innovations:

\begin{figure*}[t]
  \centering
   \includegraphics[width=1\linewidth]{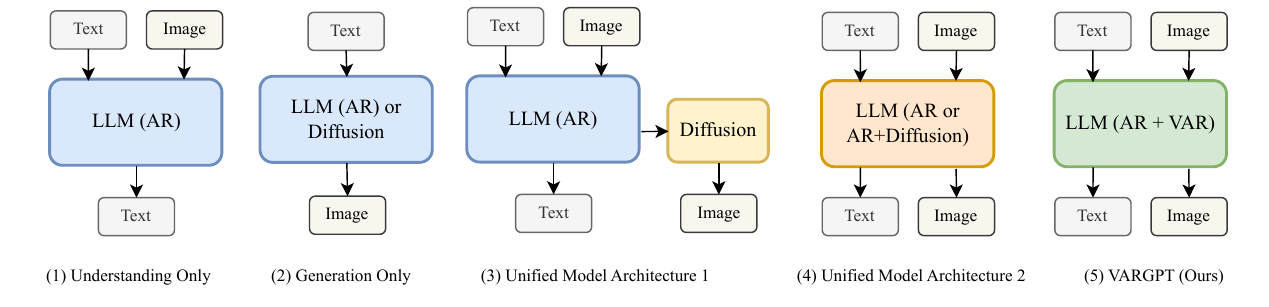}

   \caption{Comparison of different model architectures, where, `AR' denotes autoregressive, while `VAR' signifies visual autoregressive. We present a comparative analysis of architectures designed for comprehension-only tasks, generation-only tasks, and unified comprehension and generation, alongside our proposed VARGPT-v1.1 an VARGPT~\cite{ding2021cogviewmasteringtexttoimagegeneration} model. Our VARGPT-v1.1 and VARGPT are conceptualized as purely autoregressive multimodel model, achieving visual comprehension through next-token prediction and visual generation through next-scale prediction paradigms.} 
   \label{fig:vargpt_arch}
\end{figure*}

VARGPT-v1.1 incorporates four fundamental technical innovations:
(1) A multi-stage training paradigm combining iterative visual instruction tuning with reinforcement learning, where Direct Preference Optimization (DPO)~\cite{rafailov2024directpreferenceoptimizationlanguage} aligns model outputs with preferences through pairwise reward modeling. Our progressive resolution scaling (from 256$\times$256 to 512$\times$512) with alternating SFT-RL phases substantially enhances generation fidelity;
(2) An expanded visual generation corpus containing 8.3 million instruction pairs (6× larger than v1.0), comprising 4.2M carefully curated real-world samples from LAION-COCO~\cite{schuhmann2022laion5bopenlargescaledataset} and 4.1M synthetic data from Midjourney v6~\cite{pan2023journeydb} and Flux~\cite{flux2024};
(3) Enhanced visual comprehension through migration to the Qwen2-7B~\cite{qwen2} backbone, leveraging its improved tokenization efficiency and attention mechanisms;
(4) Architecture-agnostic fine-tuning using constructed image-editing datasets, enabling VARGPT-v1.1 to acquire visual editing capabilities without structural modifications.

Through these improvements, our VARGPT-v1.1 demonstrates significant advancements in both comprehension and generation capabilities, outperforming comparable unified MLLMs in multimodal tasks. Extensive experiments reveal that VARGPT-v1.1 achieves state-of-the-art visual understanding performance (as illustrated in Figure~\ref{fig:intro}) while surpassing existing unified models and multimodal large language models across various benchmarks.
Furthermore, VARGPT-v1.1 exhibits three distinctive advantages: (1) enhanced visual generation quality with improved instruction adherence~(Figure~\ref{fig:abs_generation}), (2) flexible image editing capabilities without requiring task-specific architectures, and (3) maintained generation diversity despite increased output resolution. These advancements are quantitatively validated through comprehensive human evaluations and automated metrics comparisons.

In summary, our main contributions are threefold:
\begin{itemize}
\item We propose VARGPT-v1.1, which comprehensively improves VARGPT in terms of data size, backbone model types, and training strategies, making significant progress in unifying visual understanding and generation.
\item We propose to train VARGPT-v1.1 by gradually increasing the resolution and iterating instruction tuning and reinforcement learning.
\item Extensive experiments and evaluations demonstrate that the model achieves more advanced visual understanding and image generation capabilities than the VARGPT model, and has certain image editing capabilities.
\end{itemize}

\begin{figure*}[t]
  \centering
   \includegraphics[width=0.95\linewidth]{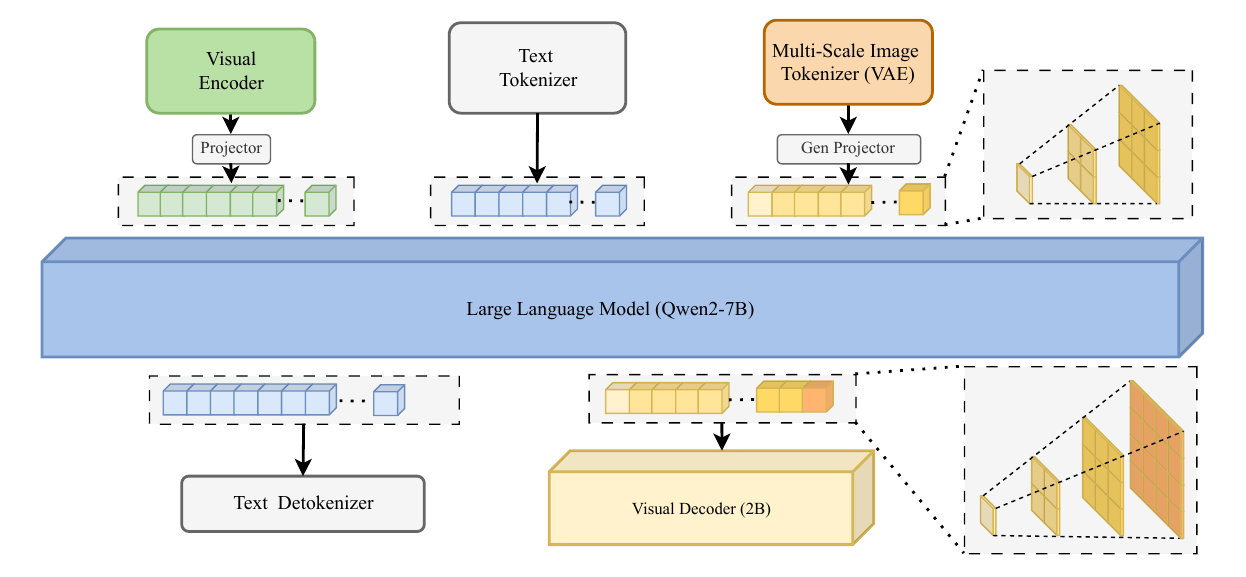}

   \caption{The illustration of the proposed VARGPT-v1.1 framework similar to VARGPT~\cite{zhuang2025vargptunifiedunderstandinggeneration}, which consists of (1) an LLM (Qwen2-7B-Instruct~\cite{Qwen2VL,qwen2}), visual encoder and a understanding projector for visual understanding; (2) a visual decoder and dual generation projectors for visual generation. VARGPT-v1.1 employs causal attention in the LLM backbone while utilizing block causal attention in the visual decoder.}
   \label{fig:main}
\end{figure*}

\section{Related Work}
\label{sec:related}

\textbf{Visual Generation and Multimodal Learning.}
Modern visual generation frameworks include diffusion models~\cite{ho2020denoisingdiffusionprobabilisticmodels, song2020generativemodelingestimatinggradients, song2022denoisingdiffusionimplicitmodels}, which reverse noise-to-image diffusion through optimized sampling~\cite{bao2022analyticdpmanalyticestimateoptimal,lu2022dpmsolverfastodesolver} and architectures~\cite{ho2021cascadeddiffusionmodelshigh,podell2023sdxlimprovinglatentdiffusion}, and flow-based models~\cite{albergo2023buildingnormalizingflowsstochastic} offering simplified alternatives. Autoregressive approaches ~\cite{esser2021tamingtransformershighresolutionimage,yu2022scalingautoregressivemodelscontentrich} adapt GPT-style token prediction, using visual tokenizers~\cite{ding2021cogviewmasteringtexttoimagegeneration, ramesh2021zeroshottexttoimagegeneration, wang2024emu3, sun2024autoregressive, sun2024autoregressivemodelbeatsdiffusion,fan2024fluid,lee2022autoregressiveimagegenerationusing} for discrete image representations. Recent scale-predictive variants (VAR~\cite{VAR}, HART~\cite{tang2024hart}) align with scaling law principles~\cite{kaplan2020scaling, henighan2020scaling}. Concurrently, Multimodal Large Language Models (MLLMs)~\cite{touvron2023llama, touvron2023llama-2} integrate visual encoders with text decoders through cross-modal connectors~\cite{li2023blip,lecun2015deep}. Systems like LLaVA~\cite{llava} and its successors~\cite{liu2024improved, liu2024llavanext,zhang2024llavanext-video,li2024llavanext-strong,li2024llavanext-ablations,li2024llavanext-interleave} enhance visual-language understanding via instructional tuning and high-quality datasets, while architectures such as Qwen-VL~\cite{bai2023qwen} and InstructBLIP~\cite{dai2023instructblipgeneralpurposevisionlanguagemodels} demonstrate advances in training paradigms and multimodal fusion. Our work unifies autoregressive generation through scale prediction while leveraging insights from these multimodal frameworks.

\textbf{Unified Models For Visual Understanding and Generation. }
In recent years, researchers have been dedicated to unifying understanding and generation capabilities within a model~\cite{ye2024xvilacrossmodalityalignmentlarge,dong2024dreamllmsynergisticmultimodalcomprehension,tang2023anytoanygenerationcomposablediffusion}. Most existing approaches~\cite{sun2024emugenerativepretrainingmultimodality,wu2023next,ge2023makingllamadrawseed} attempt to integrate pre-trained diffusion models with existing systems. However, these systems essentially treat diffusion models as external tools rather than incorporating them as intrinsic generative capabilities of MLLMs. Show-o~\cite{xie2024showosingletransformerunify} by combining autoregressive and (discrete) diffusion modeling, can adaptively handle inputs and outputs of various mixed modalities. \citet{li2024dualdiffusionunifiedimage} employs a cross-modal maximum likelihood estimation framework, significantly improving upon existing diffusion-based multimodal models. ~\cite{geminiteam2024geminifamilyhighlycapable,bai2023sequentialmodelingenablesscalable} have explored integrating image generation into large language models (LLMs) using autoregressive methods, achieving remarkable results. For instance, LWM~\cite{liu2023world} and Chameleon~\cite{chameleonteam2024chameleonmixedmodalearlyfusionfoundation} utilize a VQ tokenizer~\cite{esser2021tamingtransformershighresolutionimage,VAR} to encode images, enabling simultaneous support for multimodal understanding and generation. Janus~\cite{wu2024janusdecouplingvisualencoding} further enhances model flexibility and performance by decoupling visual encoding into separate pathways, while Dual Diffusion~\cite{li2024dualdiffusionunifiedimage} investigated the use of two diffusion models for understanding and generation. Liquid~\cite{wu2024liquid} learns images and text embeddings within the same space and implements autoregressive visual understanding and generation using the paradigm of predicting the next token.
Building on these foundational efforts, several recent works have further explored the integration of visual understanding and generation. QLIP~\cite{zhao2025qliptextalignedvisualtokenization} proposes a text-aligned visual tokenization approach that unifies auto-regressive multimodal understanding and generation, highlighting the importance of aligning visual and textual representations. VILA-U~\cite{wu2025vilauunifiedfoundationmodel} aims to develop a unified foundation model that integrates visual understanding and generation, focusing on improving the coherence between these two tasks. UniTok~\cite{ma2025unitokunifiedtokenizervisual} introduces a unified tokenizer for visual generation and understanding, demonstrating the potential of a single tokenization framework for both tasks. MINT~\cite{wang2025mintmultimodalchainthought} explores multi-modal chain of thought in unified generative models, enhancing image generation through more sophisticated reasoning mechanisms. Janus-Pro~\cite{chen2025janusprounifiedmultimodalunderstanding} scales up the data and model size for unified multimodal understanding and generation, showing that larger models and datasets can significantly improve performance.
Different from all existing unified models, we propose modeling understanding and generation as two distinct paradigms within a unified model: {predicting the next token for visual understanding and predicting the next scale for visual generation}.

\textbf{Reinforcement Learning}
Reinforcement learning (RL) has become a powerful technique to enhance the alignment, reasoning, and decision-making capabilities of large language models (LLMs). Recent works have explored using AI feedback, step-level verification, and policy optimization to improve model performance while reducing training costs. DeepSeek-R1~\cite{deepseekai2025deepseekr1incentivizingreasoningcapability} significantly advances RL for LLMs by scaling reinforcement fine-tuning (RFT) with AI feedback, achieving high-quality preference alignment without human annotations. Math-Shepherd~\cite{wang2024mathshepherdverifyreinforcellms} and DeepSeek-Math~\cite{shao2024deepseekmathpushinglimitsmathematical} introduce step-level verification and group preference optimization to improve mathematical reasoning. Notably, Group Relative Policy Optimization (GRPO)~\cite{shao2024deepseekmathpushinglimitsmathematical} eliminates the need for a critic model in policy optimization, greatly reducing training overhead while maintaining superior performance. In parallel, LIMR~\cite{li2025limrrlscaling} improves RL efficiency by training on high-impact samples, reducing data requirements by 84\%. Additionally, works like Visual-RFT~\cite{liu2025visualrftvisualreinforcementfinetuning} and Grounding-LLM~\cite{carta2024groundinglargelanguagemodels} extend RL to multimodal and interactive environments, improving decision-making and real-time response. MM-RLHF~\cite{zhang2025mmrlhfstepforwardmultimodal} introduces a dataset containing 120,000 fine-grained, human-annotated preference comparison pairs, representing a substantial advancement over existing resources in size, diversity, annotation granularity, and quality. These advances demonstrate the critical role of RL in scaling and optimizing LLM capabilities.

\section{Methodology}
\label{sec:methods}

\begin{figure*}[t]
  \centering
   \includegraphics[width=1\linewidth]{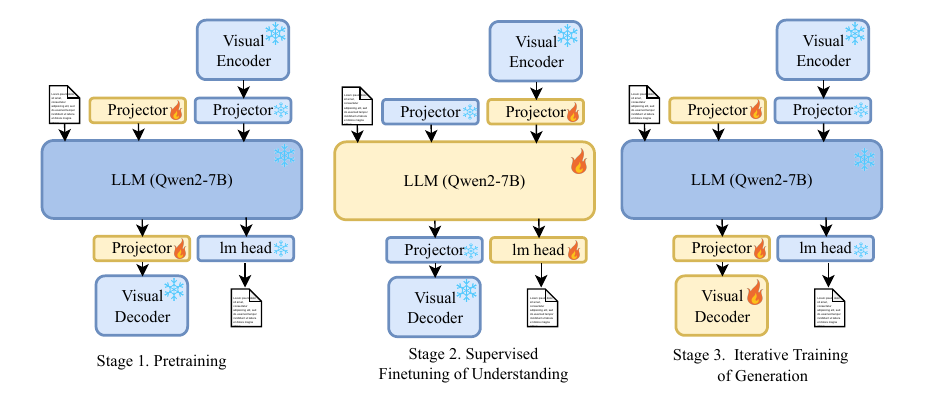}

   \caption{The three training stages of the VARGPT, including stage-1 pretraining, stage-2 visual instruction tuning and stage-3 iterative training.}
   \label{fig:vargpt_training}
\end{figure*}

\begin{figure*}[t]
  \centering
   \includegraphics[width=1\linewidth]{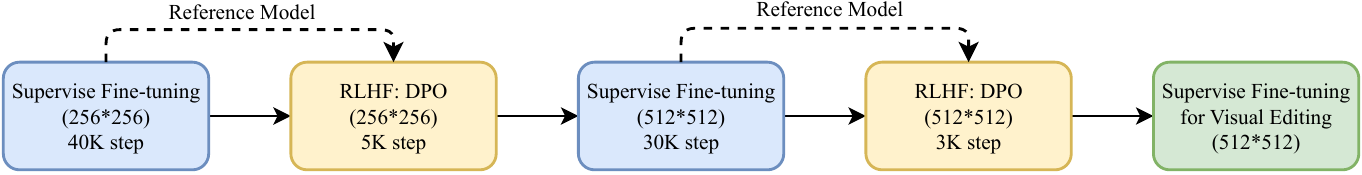}

   \caption{The proposed iterative training strategy for the third stage gradually increases the resolution of the image, while using instruction fine-tuning and reinforcement learning iterative training. Finally, we use the instruction-follow dataset for image editing to stimulate the model's visual editing ability.}
   \label{fig:vargpt_training_stage3}
\end{figure*}

\subsection{Model Architecture}
Our VARGPT-v1.1 follows the model architecture design of VARGPT to unify visual understanding and generation, which architecture is shown in the Figure~\ref{fig:main}. Our VARGPT-V1.1 architecture follows the paradigm of predicting the next token for visual understanding and adheres to the paradigm of predicting the next scale for image generation.

\textbf{Visual understanding via next-token prediction}~

Following VARGPT, our model architecture of VARGPT-v1.1 references the structure of Qwen2-VL~\cite{Qwen2VL}, using Qwen2~\cite{qwen2} as the LLM $\theta$, and employing a Vision Transformer (ViT)~\cite{radford2021learningtransferablevisualmodels} as the visual encoder along with a two-layer linear network as the projector.
Initially, the image $\mathbf{X}^{img}$ for visual understanding undergoes processing through the vision encoder to produce embeddings $\mathbf{H}^{img}$, which are then modified by the interface to align with the textual embeddings $\mathbf{H}^{txt}$ obtained by query $\mathbf{X}^{query}$. The combined data serves as input to the LLM, which autoregressively generates the textual output $\mathbf{Y}^{txt}$ as:
\begin{equation}
    \mathbf{Y}^{txt}_t \sim p_{\theta}(\mathbf{Y}^{txt}_t \mid \mathbf{X}^{img}, \mathbf{X}^{query}, \mathbf{Y}^{txt}_{<t}),
\end{equation}
where $\mathbf{Y}^{txt}_t$ represents the $t$-th token of $\mathbf{Y}^{txt}$, and $\mathbf{Y}^{txt}_{<t}$ refers to the sequence of tokens generated prior to the $t$-th step.
To maintain the causal attention property of LLM, we apply a causal attention mask to all input LLM tokens, including those used for generating images.

\textbf{Visual generation via next-scale prediction}~
For visual generation, we follow most of the settings in Infinity~\cite{Infinity} and VAR~\cite{VAR} and adopt a multi-scale image tokenizer with bitwise multi-scale residual quantizer~\cite{Infinity} for visual token encoding and decoding. We construct two image generation projectors to transform the visual features used for generation at the input and output of the LLM. Additionally, we construct an extra 2B visual decoder $\phi$ with 32 layers of Transformers to decode visual features, which can avoid conflicts between knowledge in the text decoder and image generation knowledge. The image features obtained through the visual decoder will be further decoded through the multi-scale VAE decoder to produce usable images. Unlike the text decoder (i.e., the LLM), the visual decoder uses attention that follows the block causal attention in  Infinity~\cite{Infinity} and VAR~\cite{VAR} to support predicting tokens for the next scale.
Unlike VARGPT, our VARGPT-v1.1 utilizes the infinite vocabulary classifier~\cite{Infinity} technology from Infinity to compute training losses related to visual generation.

\textbf{Prompt template for mixed-modal generation and Classifier-free guidance (CFG)}~
To differentiate between tokens designated for textual generation and those for image synthesis, we use some special token markers similar to VARGPT to train our VARGPT-v1.1. 
In addition, we use CFG techniques similar to VARGPT to improve the quality of image generation.

\subsection{Training}

The training of our VARGPT-v1.1 follows the three-stage training method of VARGPT, and the overall training process is shown in Figure~\ref{fig:vargpt_training}. However, in the stage-3, we propose the methods of iterative instruction tuning and reinforcement learning to enhance the visual generation ability of the previous model. Specifically, the iterative training process in our third stage is as shown in Figure~\ref{fig:vargpt_training_stage3}.

\subsubsection{Visual Instruction Tuning }

The instruction tuning for visual generation mainly aims to endow VARGPT-v1.1 with the visual generation ability through supervised fine-tuning. At this stage, we unfreeze the visual decoder and the two projectors and freeze other parameters to perform SFT, as shown in Figure~\ref{fig:vargpt_training_stage3}.
We employ a training method that gradually increases the image resolution to train VARGPT-v1.1. Specifically, in the first SFT stage, the image resolution is set to 256$\times$256, and the model is trained for 40,000 steps to endow it with the initial ability to generate images. In the second SFT stage, the image resolution is set to 512$\times$512, and the model is trained for 30,000 steps to further enhance its high-resolution visual generation ability.
The training data for the third stage includes 8.3M instruction pairs that are collected and constructed (for details, please refer to Section \ref{sec:dataset}).

\subsubsection{Reinforcement Learning Human Feedback}
We propose an iterative training framework that integrates instruction fine-tuning and reinforcement learning in visual autoregressive large visual-language models.
We formulate the improvement of generation quality as a preference selection problem through preference learning. This approach incentivizes the model to favor higher-quality image outputs $y_w$ while rejecting inferior counterparts $y_l$. During the preference learning phase, a reward model denoted as $\hat{r}$ is trained to reinforce the model's propensity towards producing $y_w$ outputs. This reward model demonstrates the capability to assign differential scores to various responses, thereby accurately reflecting human preferences. Subsequently, the acquired reward model $r$ is employed to provide feedback for guiding an additional fine-tuning stage, which learns a policy model $\pi_\theta$ governed by human preference guidance.

Specifically, after each SFT phase, we implement direct preference optimization (DPO)~\cite{rafailov2024directpreferenceoptimizationlanguage} for images at various resolutions to enhance the model's visual generation capabilities. For preference data construction, we sample from models at different SFT checkpoints and from the fully SFT-trained model. We sample 100k prompts from the Midjourney prompt dataset\footnote{\url{huggingface.co/datasets/vivym/midjourney-prompts}} as input prompts $x$ to obtain our model's outputs $y_l$. Concurrently, we utilize a robust commercial model (Midjourney v6 and Flux-1.1 pro) to generate images $y_w$ from these prompts. This sampling approach from both stronger and weaker models naturally creates a partially ordered dataset $\mathcal{D}=\{{(x, y_l, y_w)}\}$.
We leverage the post-SFT model as the reference model $\pi_{ref}$, and optimize the policy model using the following objective function:

\begin{equation}
\begin{aligned}
\mathcal{L}_{rl} &(\pi_\theta;\pi_{ref}) =  -\mathbb{E}_{(x, y_w, y_l)\sim\mathcal{D}} \\
&[\log \sigma(\beta\log\frac{\pi^{img}_\theta(y_w|x)}{\pi^{img}_{ref}(y_w|x)}\
-\beta\log\frac{\pi^{img}_\theta(y_l|x)}{\pi^{img}_{ref}(y_l|x)})],
\end{aligned}
\label{eq:dpo}
\end{equation}
where, $\pi^{img}_\theta$ and $\pi^{img}_{ref}$ denote the policy model and reference model for the generated image tokens, $\log \sigma$ representing the log sigmoid function and $\beta$ is the hyperparameters related to rewards.
Rather than explicitly training the reward model, Eq.~\ref{eq:dpo} directly optimizes the policy model by implicitly representing the reward $r$ through:
\begin{equation}
    \hat{r}(x, y) = \beta\log \frac{\pi^{img}_\theta(y|x)}{\pi^{img}_{ref}(y|x)}.
\end{equation}
Therefore, we can formulate Eq.~\ref{eq:dpo} to maximize the reward margin $\hat{r}(x, y_w)-\hat{r}(x, y_l)$.

The proposed VARGPT-v1.1 is capable of simultaneously generating both text tokens and image tokens. During the reinforcement learning phase, exclusive focus is placed on the generated image tokens, where the image-specific policy models $\pi^{img}_\theta$ and $\pi^{img}_{ref}$ are formulated as:
\begin{equation}
    \pi^{img}_* = \operatorname{softmax}(\operatorname{logit}_*[img]), *\in\{\theta, ref\},
\end{equation}
where, $\operatorname{logit}_*[img]$ denotes the logits corresponding to the positions of generated image tokens.

\subsubsection{SFT for Visual Editing}
Following multi-stage resolution-progressive iterations of SFT and DPO, we systematically construct an instruction-tuning dataset comprising 11,325 samples from StyleBooth~\cite{han2024styleboothimagestyleediting}~\footnote{\url{https://huggingface.co/datasets/yzwang/X2I-mm-instruction}} to enable visual editing capabilities in VARGPT-v1.1. The pipeline processes target images through the visual encoder while utilizing editing instructions as textual prompts to supervise the model's approximation of edited image distributions. Crucially, this approach achieves: (1) architecture-preserving adaptation through parameter space optimization, and (2) unified multimodal editing via joint text-image token prediction. All model parameters were unfrozen during SFT to maximize editing fidelity while maintaining generation diversity, following recent advances in full-parameter fine-tuning for multimodal systems.

\section{Instruction-following  and Preference Data}
\label{sec:dataset}

\begin{figure}
    \centering
    \begin{subfigure}{1\linewidth}
      \includegraphics[width=1\linewidth]{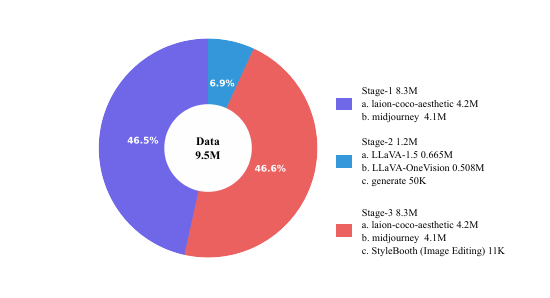}
      \label{fig:PieChart2}
    \end{subfigure}
    \caption{We present the data distribution we construct and collect, encompassing the proportional breakdown of data across the three training stages. Our composite dataset for stage-2 training is derived from LLaVA-1.5, LLaVA-OneVision.}
        \label{fig:PieChart-exp}
  \end{figure}

\begin{table*}[htbp]
  \centering
    \begin{tabular}{l|ccccccc}
    \toprule
    \multicolumn{1}{l}{\multirow{2}[4]{*}{\textbf{Hyperparameter}}} & \multirow{2}[4]{*}{\textbf{Stage 1}} & \multirow{2}[4]{*}{\textbf{Stage 2}} & \multicolumn{4}{c}{\textbf{Stage 3}} & \multirow{2}[4]{*}{\textbf{Stage e}} \\
\cmidrule{4-7}    \multicolumn{1}{l}{} &       &       & \textbf{SFT(256)} & \textbf{DPO(256)} & \textbf{SFT(512)} & \textbf{DPO(512)} &  \\
    \midrule
    batch size & 1024  & 256   & 1024  & 384   & 512   & 384   & 512 \\
    lr    & 1e-3  & 5e-5  & 5e-5  & 1e-6  & 5e-5  & 1e-6  & 5e-5 \\
    lr schedule & Cosine & Cosine & Cosine & Cosine & Cosine & Cosine & Cosine \\
    lr warmup ratio & 0.1   & 0.1   & 0.1   & 0.1   & 0.1   & 0.1   & 0.1 \\
    weight decay & 0     & 0     & 0     & 0     & 0     & 0     & 0 \\
    training step & 2k    & 3.1k  & 40k   & 5k    & 30k   & 3k    & 2k \\
    optimizer & AdamW & AdamW & AdamW & AdamW & AdamW & AdamW & AdamW \\
    DeepSpeed ZERO type & 3     & 2     & 3     & 3     & 3     & 3     & 3 \\
    \bottomrule
    \end{tabular}%
      \caption{Main hyperparameters and training settings for training our VARGPT-v1.1 at various stages, where, (256) and (512) represent the image generation resolutions of $256\times256$ and $512\times512$, respectively}
  \label{tab:setting}%
\end{table*}%

\subsection{Instruction-following Data}
For the visual understanding instruction-following dataset, we maintain complete consistency with VARGPT's training data composition from LLaVA-1.5-665K~\cite{liu2024improvedbaselinesvisualinstruction} and  LLaVA-OneVision~\cite{liu2024llavanext,zhang2024llavanext-video,li2024llavanext-strong,li2024llavanext-ablations,li2024llavanext-interleave}, and same preprocessing pipeline. The visual generation dataset comprises 8.3M image-text pairs from two complementary sources: (1) 4.2M real-world images from LAION-COCO~\cite{laion-coco-aesthetic} filtered through aesthetic scoring, and (2) 4.1M synthetic samples generated by Midjourney collected from JourneyDB~\cite{pan2023journeydb} and other public repositories. This 1:1 real-synthetic ratio leverages recent findings that synthetic data accelerates model convergence~\cite{chen2025janusprounifiedmultimodalunderstanding}.

 We implement a three-stage curriculum training strategy: For the Stage 1: Full 8.3M generation dataset for foundational capabilities. For the Stage 2: Hybrid training with 1.18M understanding samples (from VARGPT) and 50K generation samples from constructed 8.3M training data.
For the Stage 3: Final refinement using the complete 8.3M generation set.
The progressive training framework inherits VARGPT's instruction construction methodology while introducing dynamic batch balancing between modalities.

\subsection{Preference Instruction Data}
We employ iterative instruction tuning and reinforcement learning to train our VARGPT-v1.1 model, which necessitates the construction of partially ordered datasets for implementing DPO. 
We randomly sample 180k prompts from the Midjourney prompt dataset\footnote{\url{huggingface.co/datasets/vivym/midjourney-prompts}}. For each prompt, we generate "losing" images using our SFT model after initial training, while utilizing corresponding Midjourney and Flux-generated images as "winning" examples to construct the partially ordered dataset for DPO training.
This approach leverages the prior that commercial generative models currently outperform our in-training model to automatically create preference data, effectively circumventing the need for labor-intensive manual annotation.

\subsection{Image Editing Data}
Our VARGPT-v1.1 demonstrates inherent advantages in visual editing capabilities due to its dual capacities in visual understanding and generation, requiring no architectural modifications for this specific task. Our system processes visual editing tasks by feeding input images through the vision encoder while incorporating user editing instructions to produce modified outputs.

For training the visual editing capacities, we construct an instruction tuning dataset using 11,325 samples from Stylebooth~\cite{han2024styleboothimagestyleediting}\footnote{\url{https://huggingface.co/datasets/yzwang/X2I-mm-instruction}}. All images were standardized at 512$\times$512 resolution, with input images processed through our visual understanding preprocessing pipeline and output images handled through the dedicated generation preprocessing protocol.

\section{Experiments}
\label{sec:experiments}

\begin{table*}[t]
  \centering
    \begin{tabular}{c|cc|ccccccc}
    \toprule
    \multirow{2}[2]{*}{\textbf{Methods}} & \multirow{2}[2]{*}{\textbf{LLM}} & \multicolumn{1}{c|}{\multirow{2}[2]{*}{\textbf{Gen.}}} & \multirow{2}[2]{*}{\textbf{MMBench}} & \multirow{2}[2]{*}{\textbf{SEED }} & \multirow{2}[2]{*}{\textbf{MMMU}} & \multicolumn{3}{c}{\textbf{POPE (acc.)}} & \multirow{2}[2]{*}{\textbf{MME }} \\
          &       &       &       &       &       & \textbf{adv} & \textbf{pop} & \textbf{rand} &  \\
    \midrule
    Chameleon (7B) & from scratch &  $\checkmark$     & 31.1  & 30.6  & 25.4  &       &       &       & 170 \\
    SEEDLLaMA (7B) & Vicuna-7B &  $\checkmark$     & 45.8  & 51.5  & -     & -     & -     & -     & - \\
    Show-o (1.5B) & Phi-1.5-1.3B &  $\checkmark$     & -     & -     & 25.1  & -     & -     & -     & 948.4 \\
    VILA-U(7B) & LLaMA-2-7B  &  $\checkmark$     & -     & 59.0  & -     & -     & -     & -     & 1401.8 \\
    Liquid (7B) & Gemma-7B &  $\checkmark$     & -     & -  & -     & -     & -     & 81.1     & 1119.3 \\
    Janus-Pro(7B)  &  DeepSeek-LLM & $\times$     & -  & 72.1  & 41.00 & -  & - & - & 1567.1 \\
    
    \midrule
    MiniGPT-4 (7B) & Vicuna-7B &  $\times$     & 23.0  & 47.4  & \multirow{6}[1]{*}{-} & 65.17 & 69.73 & 79.67 & 581.7 \\
    InstructBLIP (8B) & Vicuna-7B & $\times$     & 36.0  & 53.4  &       & 72.10     & 82.77     & {88.57}     & 1212.8 \\
    Qwen-VL (7B) & Qwen-7B & $\times$     & 38.2  & -     &       & -     & -     & -     & - \\
    Qwen-VL-Chat (7B) & Qwen-7B & $\times$     & 60.6  & 58.2  &       & -     & -     & -     & 1487.5 \\
    LLaVA (7B) & Vicuna-7B & $\times$     & 36.2  & 33.5  &       & 49.70 & 49.87 & 50.37 & 502.8 \\
    mPLUG-Owl2 (7B)  & LLaMA-2-7B  & $\times$     & 64.5  & 57.8  &       & -     & -     & -     & 1450.2 \\
    LLaVA-1.5-hf (7B) & Vicuna-7B & $\times$     & 62.7  & 65.4  & 35.24 & 83.60  & 85.77 & 86.97 & 1480.1 \\
    \midrule
    \rowcolor{aliceblue} {VARGPT(7B+2B)} & {Vicuna-7B} &  $\checkmark$     & {67.6} & {67.9} & {36.44} & {84.40 } & {85.90 } & {87.37} & {1488.8} \\
    \rowcolor{aliceblue} \textbf{VARGPT-v1.1(7B+2B)} & \textbf{Qwen2-7B} &  $\checkmark$     & \textbf{81.01} & \textbf{76.08} & \textbf{48.56} & \textbf{84.40 } & \textbf{88.36 } & \textbf{89.17} & \textbf{1684.1} \\
    
    \bottomrule
    \end{tabular}%
  \caption{Zero-shot multi-modal evaluation on multi-modal benchmarks including MMMU, MME, MMBench, SEEDBench, and POPE (including different settings \textit{random}, \textit{popular} and \textit{adversarial} ). The overall scores are reported for evaluation and we report test results for MMBench. \textbf{Gen} represents whether the method supports image generation capability. VARGPT-v1.1 achieves the best overall
performance.}
\vspace{-0.5em}
  \label{tab:main-1}%
\end{table*}%

\begin{table*}[htbp]
  \centering
    \begin{tabular}{c|cc|cccccc}
    \toprule
    \multirow{2}[2]{*}{\textbf{Methods}} & \multirow{2}[2]{*}{\textbf{LLM}} & \multicolumn{1}{c|}{\multirow{2}[2]{*}{\textbf{Gen.}}} & \multirow{2}[2]{*}{\textbf{GQA}} & \multirow{2}[2]{*}{\textbf{TextVQA}} & \multirow{2}[2]{*}{\textbf{VQAv2}} & \multirow{2}[2]{*}{\textbf{SciQA-img}} & \multirow{2}[2]{*}{\textbf{OKVQA}} & \multirow{2}[2]{*}{\textbf{VizWizQA}} \\
          &       &       &       &       &       &       &       &  \\
    \midrule
    MiniGPT-4 (7B) & Vicuna-7B & $\times$      & 43.5  & -     & 0.6   & 39.6  & -     & - \\
    InstructBLIP (8B) & Vicuna-7B & $\times$      & 49.2  & -     & -     & 60.5  & -     & 34.5 \\
    Qwen-VL (7B) & Qwen-7B & $\times$      & 59.3  & 50.1  & -     & 67.1  & -     & 35.2 \\
    Qwen-VL-Chat (7B) & Qwen-7B & $\times$      & 57.5  & {61.5} & 78.2  & 68.2  & 56.6  & 38.9 \\
    mPLUG-Owl2 (7B)  & LLaMA-2-7B  & $\times$      & 56.1  & 53.3  & -     & 68.7  & {57.7} & 54.5 \\
    LLaVA-1.5-hf (7B) & Vicuna-7B & $\times$      & 61.2  & 48.8  & 76.49 & 67.9  & 53.2  & 53.53 \\
    \midrule
\rowcolor{aliceblue}    {VARGPT(7B+2B)} & {Vicuna-7B} & $\checkmark$ & {62.3} & {54.1} & {78.4} & {80.1} & 55.8  & {56.8} \\
\rowcolor{aliceblue}    \textbf{VARGPT-v1.1(7B+2B)} & \textbf{Qwen2-7B} & $\checkmark$ & \textbf{66.2} & \textbf{82.0} & \textbf{80.4} & \textbf{91.6} & \textbf{57.8}   & \textbf{65.4} \\
    \bottomrule
    \end{tabular}%
  \caption{ Performance comparison on visual question answering tasks.  We \textcolor{gray}{gray} out the model has trained on the dataset. \textbf{Gen} represents whether the method supports image generation capability.}
  \label{tab:main-2}%
\end{table*}%

\begin{table*}
\centering
\begin{tabular}{lcccccc}
\midrule
Method & Global$\uparrow$ & Entity$\uparrow$ & Attribute$\uparrow$ & Relation & Other & Overall$\uparrow$ \\
\midrule
SDv1.5 & 74.63 & 74.23 & 75.39 & 73.49 & 67.81 & 63.18 \\
PixArt-$\alpha$  & 74.97 & 79.32 & 78.60 & 82.57 & 76.96 & 71.11 \\
Lumina-Next & 82.82 & 88.65 & 86.44 & 80.53 & 81.82 & 74.63 \\
SDXL  & 83.27 & 82.43 & 80.91 & 86.76 & 80.41 & 74.65 \\
Playground v2.5  & 83.06 & 82.59 & 81.20 & 84.08 & 83.50 & 75.47 \\
\midrule
\rowcolor{aliceblue}  \textbf{VARGPT-v1.1(7B+2B)}  & 84.83 & 82.80 & 84.95 & 88.13 & 87.70 & \textbf{78.59} \\
\midrule
\end{tabular}
 \caption{ Performance comparison on the DPG-Bench benchmark.  }
\label{exp:gen-2}
\end{table*}

\begin{table*}
\centering
\begin{tabular}{llccccccc}
\midrule
Type & Method & Single Obj. & Two Obj. & Counting & Colors & Position & Color Attr. & Overall$\uparrow$ \\
\midrule
& LlamaGen  & 0.71 & 0.34 & 0.21 & 0.58 & 0.07 & 0.04 & 0.32 \\
& LDM  & 0.92 & 0.29 & 0.23 & 0.70 & 0.02 & 0.05 & 0.37 \\
& SDv1.5  & 0.97 & 0.38 & 0.35 & 0.76 & 0.04 & 0.06 & 0.43 \\
& PixArt-$\alpha$ & 0.98 & 0.50 & 0.44 & 0.80 & 0.08 & 0.07 & 0.48 \\
Gen. Only & SDv2.1  & 0.98 & 0.51 & 0.44 & 0.85 & 0.07 & 0.17 & 0.50 \\
& DALL-E 2  & 0.94 & 0.66 & 0.49 & 0.77 & 0.10 & 0.19 & 0.52 \\
\midrule
& SEED-X$^\dagger$ & 0.97 & 0.58 & 0.26 & 0.80 & 0.19 & 0.14 & 0.49 \\
& Show-o & 0.95 & 0.52 & 0.49 & 0.82 & 0.11 & 0.28 & \textbf{0.53} \\
Und. and Gen. & LWM  & 0.93 & 0.41 & 0.46 & 0.79 & 0.09 & 0.15 & 0.47 \\

& Chameleon  & - & - & - & - & - & - & 0.39 \\

 \rowcolor{aliceblue} & \textbf{VARGPT-v1.1} & 0.96 & 0.53 & 0.48 & 0.83 & 0.13 & 0.21 & \textbf{0.53} \\
\bottomrule
\end{tabular}
 \caption{ Performance comparison on the GenEval benchmark.  }
 \label{exp:gen-1}
\end{table*}

\begin{figure*}[t]
  \centering
   \includegraphics[width=1\linewidth]{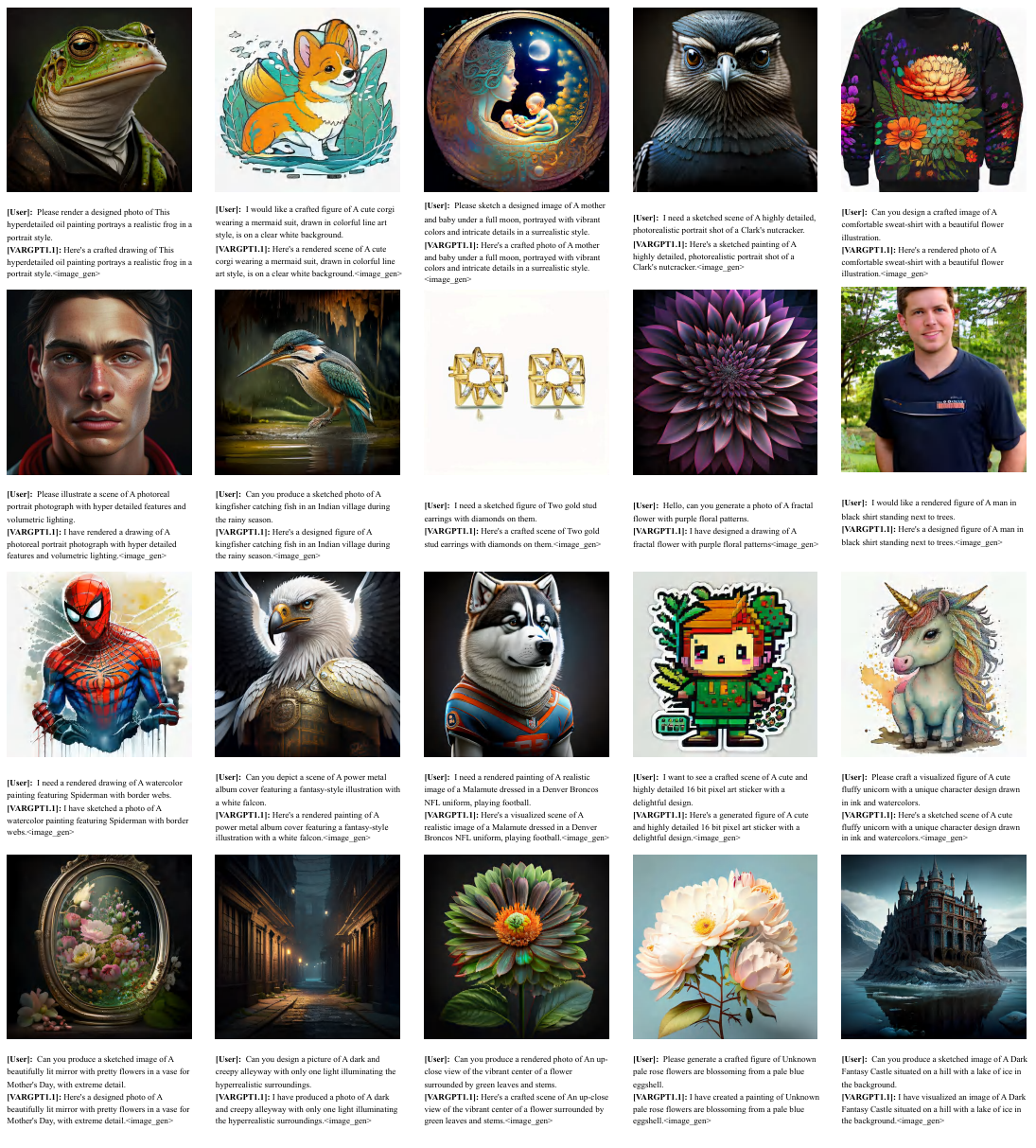}

   \caption{Some generated 512$\times$512 samples by VARGPT-v1.1. Our VARGPT-v1.1 supports text-and-image instructions from user and outputs both text-and-image mixed modal data simultaneously.}
   \label{fig:generate-vis}
\end{figure*}

\begin{figure*}[t]
  \centering
   \includegraphics[width=1\linewidth]{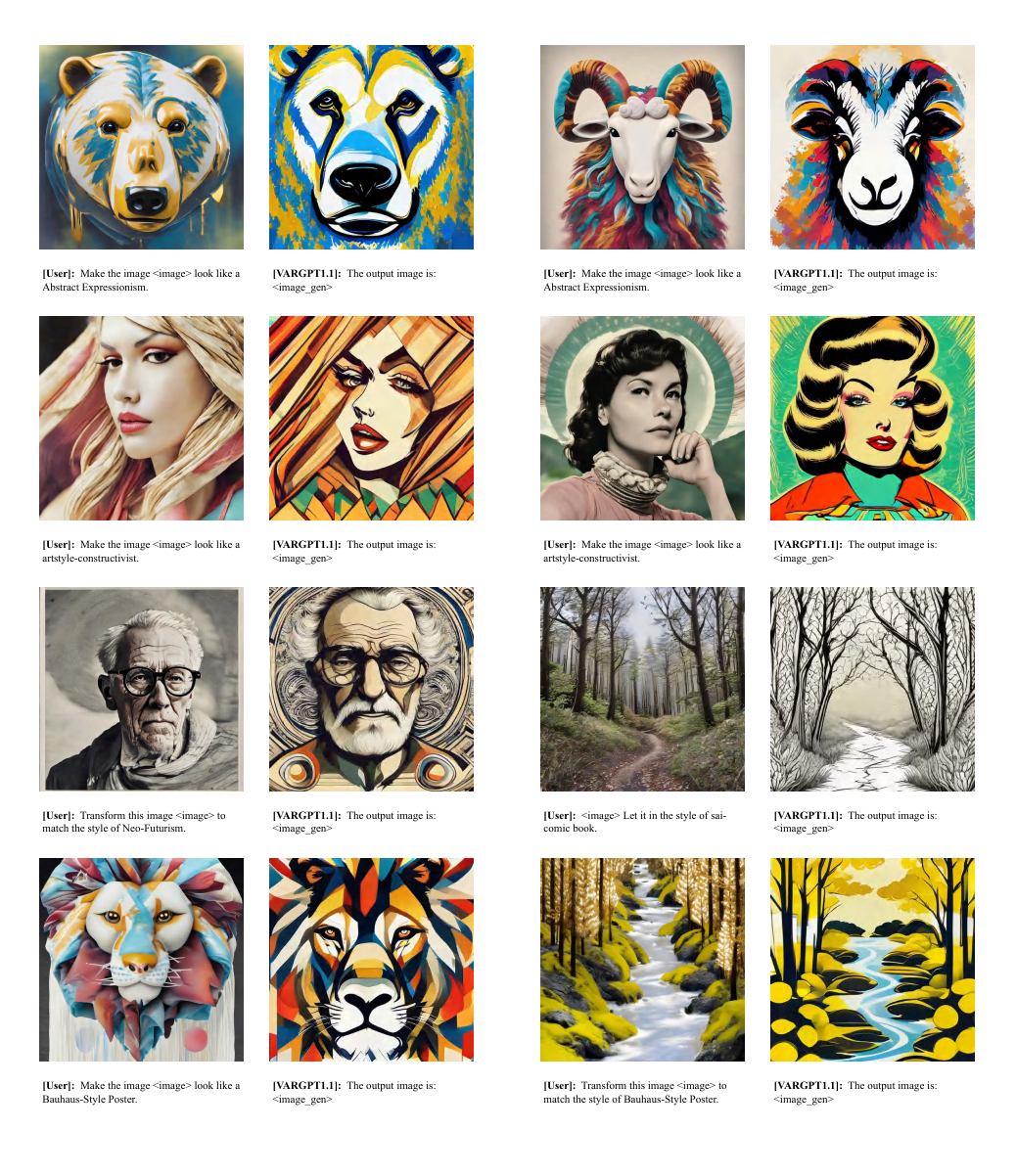}

   \caption{Some visual display of image editing results from VARGPT-v1.1, with all images having a resolution of 512$\times$512.}
   \label{fig:edit-vis}
\end{figure*}

\begin{table}
  \begin{minipage}{0.99\linewidth}
\centering
\scalebox{0.80}{
\begin{tabular}{l p{7.5cm} }
\toprule
 \multicolumn{2}{l}{\bf Visual input example:}  \\
\midrule
&  \includegraphics[height=3cm]{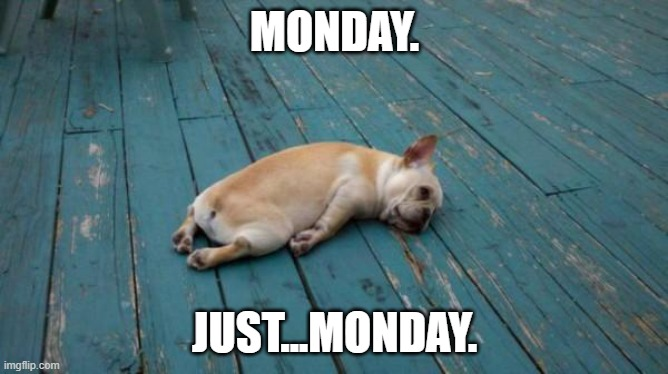} \\

User & Please explain the meme in detail. \\
\midrule
LLaVA-v1.5-7B & The image features a small brown dog lying on the floor, possibly on a wooden surface, with its head up and eyes open. The dog appears to be sleeping or resting, and it is positioned in a way that it seems to be looking up at the camera. The scene is set against a blue background, which adds a contrasting color to the image. The image is accompanied by a caption that reads "Just Monday...", suggesting that the dog is enjoying a lazy day at home. The caption is placed at the bottom of the image, with the dog taking up most of the frame. \\
\midrule
VARGPT-v1.1 &The meme features a small dog lying on its back on a wooden surface, with the words "MONDAY" and "JUST...MONDAY" written above and below it, respectively. The text humorously implies that the dog is expressing a common sentiment of fatigue or disinterest associated with Mondays, a day often associated with the start of the workweek.\\
\bottomrule
\end{tabular}
}
\captionof{table}{VARGPT demonstrates the capability to comprehend and interpret humorous elements within visual content.}
\label{tab:tricky_example}  
  \end{minipage}
\end{table}

\textbf{Details.}~
For images utilized in generation tasks, we uniformly resize and crop them to 256$\times$256 and 512$\times$512 pixels, subsequently applying preprocessing techniques consistent with those employed in infinity~\cite{Infinity, VAR}.
Regarding images for visual comprehension tasks, we adhere to the preprocessing protocol established in the Qwen2-VL~\cite{Qwen2VL} framework.
Our language model, visual encoder, and visual feature mapper are initialized using the Qwen2VL-7B-Instruct~\footnote{\url{https://huggingface.co/Qwen/Qwen2-VL-7B-Instruct}} architecture. The visual decoder is initialized with Infinity-2B~\cite{Infinity} parameters, encompassing approximately 2 billion model parameters. The visual generation feature mapper in VARGPT-v1.1 undergoes random initialization and is preliminarily updated during the first stage of pre-training.
We adopt a multi-scale VQVAE~\cite{esser2021tamingtransformershighresolutionimage} similar to Infinity~\cite{Infinity,VAR} for image tokenization, facilitating the next-scale prediction paradigm. A comprehensive summary of the training details across the three stages of our model is presented in Table~\ref{tab:setting}.
During image generation, the top-k and top-p sampling parameters of our VARGPT-v1.1 model are set to 900 and 0.95, respectively. Additionally, the classifier-free guidance (CFG) scale parameter is configured at 1.5.

\textbf{Benchmarks.}~
Following common settings of VARGPT~\cite{liu2024llavanext, zhu2023minigpt, liu2024improvedbaselinesvisualinstruction}, 
we evaluate the effectiveness of our VARGPT-v1.1 in visual understanding on a collection of both academic task-oriented benchmarks and recent benchmarks specifically proposed for instruction-following MLLMs, totaling 11 benchmarks:

(1) five multi-modal benchmarks for instruction-following MLLMs, including MMbench-dev (en)~\cite{liu2024mmbenchmultimodalmodelallaround}, SEED-bench~\cite{li2023seedbenchbenchmarkingmultimodalllms}, MMMU~\cite{yue2024mmmumassivemultidisciplinemultimodal}, POPE~\cite{li-etal-2023-evaluating} and MME~\cite{fu2024mmecomprehensiveevaluationbenchmark} benchmarks. For the POPE benchmark, we evaluate on \textit{random}, \textit{popular} and \textit{adversarial} settings with accuracy as metrics;
(2) six visual-centric question-answer benchmarks, including GQA~\cite{hudson2019gqanewdatasetrealworld}, TextVQA~\cite{singh2019vqamodelsread}, VQAv2~\cite{goyal2017makingvvqamatter}, SciQA-img~\cite{lu2022learnexplainmultimodalreasoning}, OKVQA~\cite{marino2019okvqavisualquestionanswering} and VizWizQA~\cite{gurari2018vizwizgrandchallengeanswering}. 
For visual understanding benchmarks, we use the settings in lmms-eval~\cite{zhang2024lmmsevalrealitycheckevaluation} to achieve a unified evaluation.

For visual generation assessment, we adopt existing evaluation benchmarks for generative models, utilizing GenEval~\cite{ghosh2023genevalobjectfocusedframeworkevaluating} and DPG-bench~\cite{hu2024ellaequipdiffusionmodels}~\footnote{\url{ https://github.com/TencentQQGYLab/ELLA }} as the evaluation frameworks for assessing image generation capabilities. We employ the original prompts provided in these two benchmarks for evaluation purposes, strictly following all settings specified in the original benchmarks.
For evaluations using the GenEval benchmark, we implement assessments following the prompt rewriting setup described in Infinity~\cite{Infinity}.

\textbf{Baselines.}~
For visual understanding, we conducted a comparative analysis of our VARGPT-v1.1 model against other multimodal large language models of similar scale. The comparison included prominent models such as LLaVA-1.5~\cite{Liu2023VisualIT}, MiniGPT-4~\cite{Chen2023MiniGPTv2LL}, and mPLUG-Owl2~\cite{Ye2023mPLUGOwl2RM}, as well as InstructBLIP~\cite{dai2023instructblipgeneralpurposevisionlanguagemodels} and Qwen-VL~\cite{bai2023qwen}.
Furthermore, we extended our comparative study to encompass unified models, including Chameleon~\cite{chameleonteam2024chameleonmixedmodalearlyfusionfoundation}, SEEDLLaMA~\cite{ge2023makingllamadrawseed}, Show-o~\cite{xie2024showo}, and VILA-U~\cite{xie2024showo}. This comprehensive comparison allowed us to evaluate the performance of VARGPT-v1.1 in relation to a diverse range of state-of-the-art models in the field.
For visual generation, 
we compare VARGPT-v1.1 with existing pure generative models including LlamaGen~\cite{sun2024autoregressivemodelbeatsdiffusion}, LDM~\cite{rombach2022highresolutionimagesynthesislatent}, PixArt-$\alpha$~\cite{chen2023pixartalphafasttrainingdiffusion}, SDv1.5~\cite{rombach2022highresolutionimagesynthesislatent}, SDv2.1~\cite{rombach2022highresolutionimagesynthesislatent}, DALL-E2~\cite{ramesh2022hierarchicaltextconditionalimagegeneration} , Playground v2.5~\cite{li2024playgroundv25insightsenhancing}, SDXL~\cite{podell2023sdxlimprovinglatentdiffusion} and Lumina-Next~\cite{zhuo2024luminanextmakingluminat2xstronger}, as well as with state-of-the-art unified models such as Seed-X~\cite{ge2024seed}, Show-o~\cite{xie2024showo}, Chameleon~\cite{chameleonteam2024chameleonmixedmodalearlyfusionfoundation}, LWM~\cite{liu2025worldmodelmillionlengthvideo}.

\subsection{Main Results}

\noindent\textbf{Evaluation on Multi-modal Benchmarks.}~
We conduct comprehensive evaluations of VARGPT-v1.1 against various state-of-the-art multimodal models, with the results presented in Table~\ref{tab:main-1}.
Based on the experimental results, we have:
(1) VARGPT-v1.1 demonstrates superior performance across all benchmarks, achieving state-of-the-art results with 81.01 on MMBench, 76.08 on SEED, and 48.56 on MMMU, surpassing previous models including LLaVA-1.5~\cite{Liu2023VisualIT}, InstructBLIP~\cite{dai2023instructblipgeneralpurposevisionlanguagemodels} and Janus-pro~\cite{chen2025janusprounifiedmultimodalunderstanding}.
(2) In comparison to generation-capable models such as Chameleon~\cite{chameleonteam2024chameleonmixedmodalearlyfusionfoundation} and SEEDLLaMA~\cite{ge2023makingllamadrawseed}, VARGPT-v1.1 shows remarkable improvements across multiple metrics while maintaining generation capabilities. 
Furthermore, the consistent performance improvements on the LLaVA-Bench benchmark, with partial results presented in Table~\ref{tab:tricky_example}, validate the effectiveness of our architectural choices and training strategy, establishing VARGPT-v1.1 as a robust and versatile multimodal model.

\noindent\textbf{Evaluation on Visual Question Answering Benchmarks.}~
We evaluate VARGPT-v1.1's performance across multiple visual question answering datasets, comparing it with several state-of-the-art multimodal models. The results are presented in Table~\ref{tab:main-2}.
Our experimental results demonstrate several notable findings:
(1) VARGPT-v1.1 achieves superior performance across all VQA benchmarks, showing significant improvements over existing models. Specifically, it achieves 66.2\% on GQA, 82.0\% on TextVQA, 80.4\% on VQAv2, and a remarkable 91.6\% on SciQA-img, establishing new state-of-the-art results on these benchmarks.
(2) Compared to its predecessor VARGPT (7B+2B), the v1.1 version shows substantial improvements across all metrics, with particularly notable gains in TextVQA (54.1\% to 82.0\%) and SciQA-img (80.1\% to 91.6\%), demonstrating the effectiveness of our architectural improvements and training strategy.
These comprehensive results validate VARGPT v1.1's superior capabilities in visual-language understanding across diverse question-answering scenarios.

\noindent\textbf{Evaluation on Instruction-to-image Task.}~

To evaluate the visual generation capabilities of VARGPT, we conducted comprehensive assessments using the widely adopted GenEval benchmark and DPG-Bench benchmark, with the quantitative results presented in Table~\ref{exp:gen-1} and Table~\ref{exp:gen-2} respectively. These datasets provide rigorous evaluation frameworks for text-to-image generation capabilities.
Our experimental results demonstrate that VARGPT-v1.1 outperforms numerous specialized image-generation models, including diffusion-based architectures (e.g., SDv2.1) and autoregressive approaches (e.g., LlamaGen). Furthermore, when compared with existing unified model baselines, our approach achieves significant improvements in accurate text-to-image generation.
As visualized in Figure~\ref{fig:generate-vis}, we present representative image outputs and conversational interactions generated by VARGPT-v1.1. Qualitative analysis reveals that VARGPT-v1.1 consistently produces high-quality images that are closely aligned with given textual instructions.

Notably, VARGPT-v1.1 maintains the original VARGPT architecture's capability to seamlessly integrate textual descriptions and visual generation within a single conversational flow. 
Compared with the image generation datasets used by other unified models (Show-o: 36M parameters, VILA-U: 15M parameters, Liquid: 30M image parameters), VARGPT-v1.1 achieves superior image generation performance with significantly fewer parameters (8.3M). This remarkable efficiency demonstrates VARGPT-v1.1's scalability in achieving high-quality visual generation, along with its unique advantages in unifying visual understanding and generation tasks within a single framework.

\noindent\textbf{Visualization of Visual Editing Results.}~
Qualitative evaluation of our visual editing capabilities, as illustrated in Figure\ref{fig:edit-vis}, demonstrates that VARGPT-v1.1 achieves fundamental image manipulation competence. This capability emerges solely through training with visual editing instruction fine-tuning data, without requiring any architectural modifications. Furthermore, these observations substantiate the substantial potential of our unified model architecture in achieving generalized visual understanding, generation, and editing within a single framework.

\section{Conclusion, Limitation and Future Work}
\label{sec:conclusion}

\textbf{Conclusion.}~
This work presents VARGPT-v1.1, an advanced extension of our unified visual autoregressive framework VARGPT, which achieves breakthrough performance in visual understanding and generation tasks through synergistic optimization of multi-stage training paradigms and large-scale data utilization. The model innovatively integrates iterative visual instruction tuning with DPO-based reinforcement learning, coupled with progressive resolution scaling strategies that significantly enhance generation fidelity. Notably, the framework realizes image editing capabilities through instruction tuning without structural modifications. These advancements demonstrate VARGPT-v1.1's scalability through flexible adoption of training strategies originally developed for MLLMs, while establishing new technical pathways for multimodal system architecture design.

\textbf{Limitation.}~
\label{sec:limitation}
(1) There remains a discernible quality gap between VARGPT-v1.1 and certain commercial generative models, such as more advanced models like FLUX~\cite{flux2023}, which are pre-trained on extensive high-quality image datasets. This disparity is primarily attributable to differences in training data.  
Scaling on larger datasets and models can alleviate this difference.
(2)  Although VARGPT-v1.1 has shown preliminary performance in visual editing, due to the limited nature of the visual editing dataset, some subtle details in the instructions may not be fully reflected in the generated images and can be used for more complex editing tasks.
(3) Since the data used to train our model's visual editing capabilities comes exclusively from Stylebooth~\cite{han2024styleboothimagestyleediting}, the current editing capabilities are thus limited to style transfer-type visual edits.

\textbf{Future Work.}~
\label{sec:Future_Work}
(1) To achieve superior image generation quality and enable higher-resolution outputs, our technical roadmap includes three strategic initiatives: exploration of novel tokenizer architectures, systematic scaling of model parameters and training datasets, and implementation of large-scale hybrid training paradigms. This multi-dimensional approach aims to establish new quality benchmarks while maintaining computational efficiency.
(2) For subsequent iterations of VARGPT, we plan to investigate reinforcement learning integration within the autoregressive framework. This research direction specifically targets enhancing generation consistency in multi-turn dialogues and improved instruction adherence through reward modeling techniques.
\maketitlesupplementary

\section{Qualitative Study for Understanding}

To further evaluate our method's effectiveness in visual understanding similar to VARGPT, we conducted experiments on LLaVA-Bench~\cite{Liu2023ImprovedBW}, which consists of 24 distinct images with expert-annotated descriptions and corresponding evaluation questions. 
In alignment with previous studies~\cite{Yin2023WoodpeckerHC, Leng2023MitigatingOH, Chen2024HALCOH}, we employed this benchmark for qualitative assessment. 
The visual results are presented in Table~\ref{tab:tricky_example-1}, ~\ref{tab:tricky_example-2}, ~\ref{tab:tricky_example-3},~\ref{tab:tricky_example-4}, and \ref{tab:tricky_example-5}.

\section{Qualitative Study for Generation}
\begin{figure*}[h]
  \centering
   \includegraphics[width=1\linewidth]{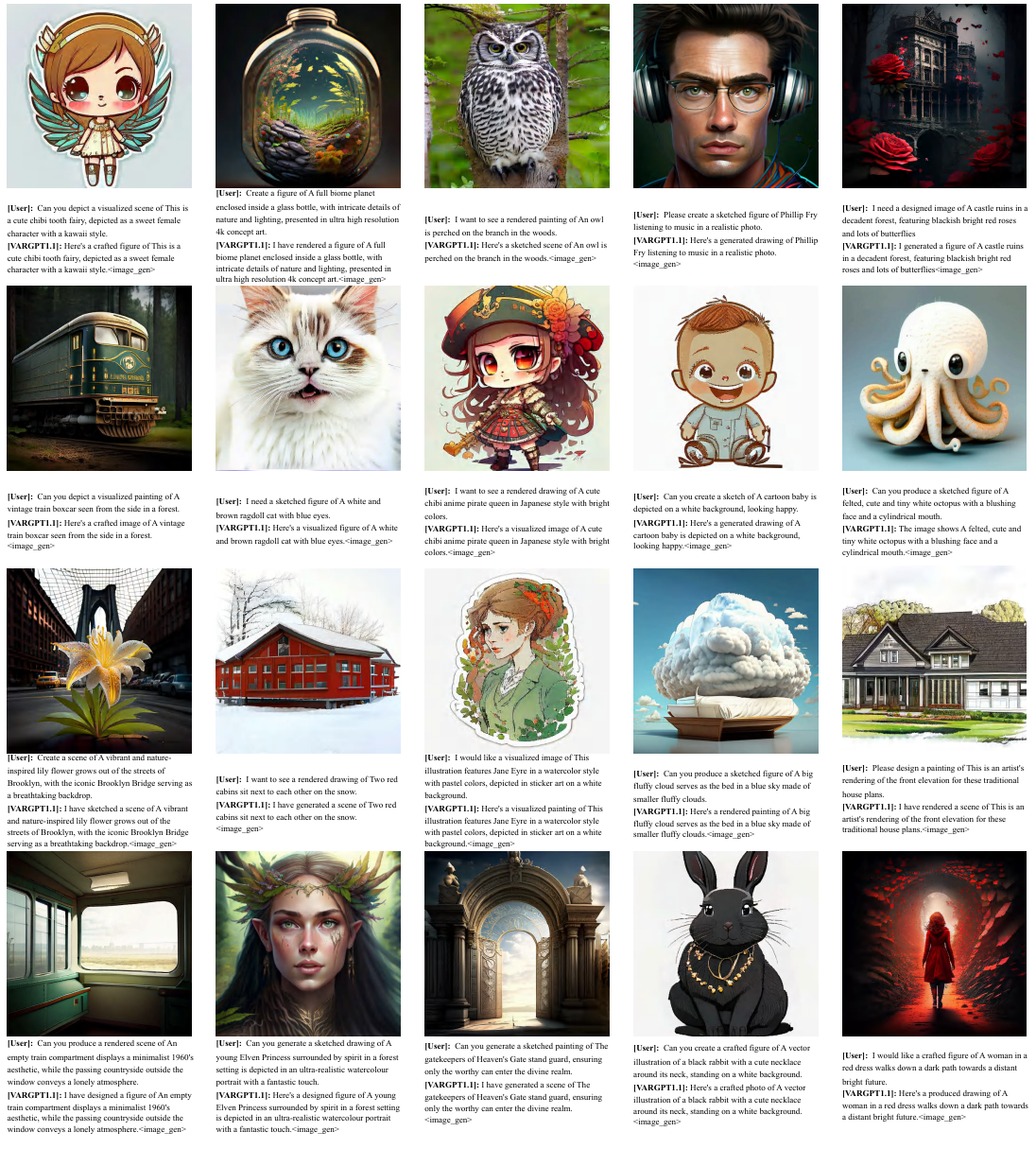}

   \caption{Some generated 512$\times$512 samples by VARGPT-v1.1. Our VARGPT-v1.1 supports text-and-image instructions from user and outputs both text-and-image mixed modal data simultaneously.}
   \label{fig:instruction-following-generate-1}
\end{figure*}

\begin{figure*}[h]
  \centering
   \includegraphics[width=0.9\linewidth]{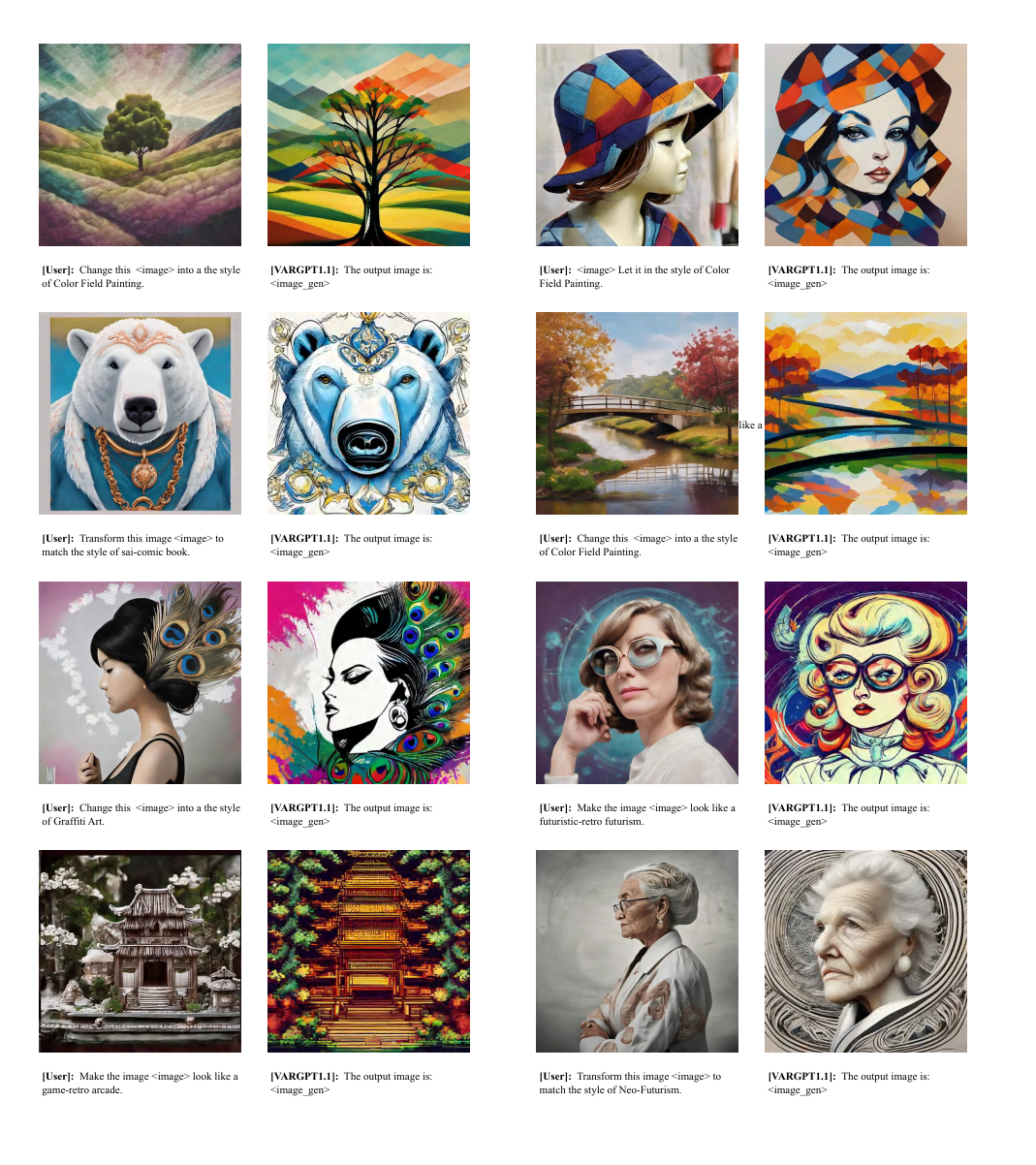}

   \caption{Some visual display of image editing results from VARGPT-v1.1, with all images having a resolution of 512$\times$512.}
   \label{fig:instruction-following-generate-2}
\end{figure*}

In Figures~\ref{fig:instruction-following-generate-1} and~\ref{fig:instruction-following-generate-2}, we present additional results demonstrating our VARGPT-v1.1's capability to output mixed-modality data from various datasets similar to VARGPT, as well as a quantitative assessment of the quality of instruction-generated images. We observe that our approach naturally supports mixed-modal data, seamlessly embedding visual generation tokens within text-generated tokens, while also producing high-quality image content. These findings underscore the significant effectiveness of our VARGPT-v1.1 in unifying understanding and generation tasks.

\begin{table}
  \begin{minipage}{0.99\linewidth}
\centering
\scalebox{0.80}{
\begin{tabular}{l p{7.5cm} }
\toprule
 \multicolumn{2}{l}{\bf Visual input example:}  \\
\midrule
&  \includegraphics[height=3.5cm]{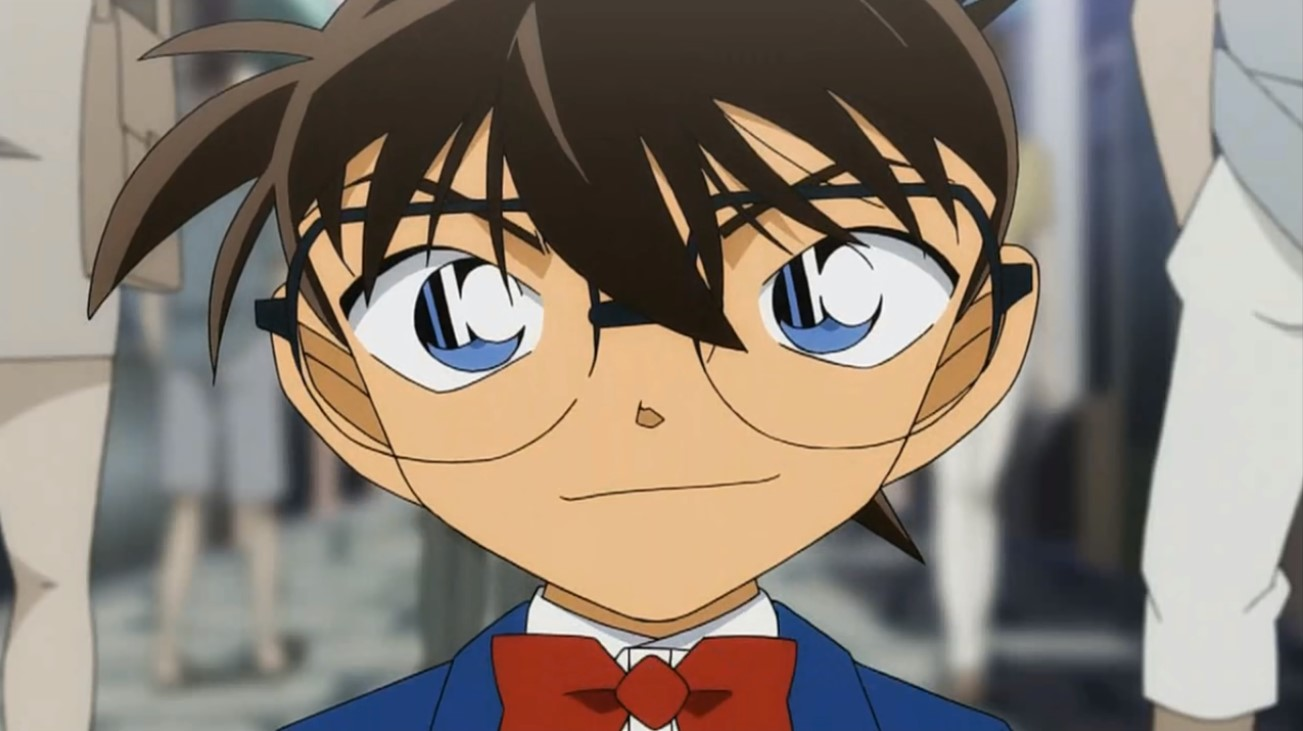} \\

User & What's the personality of this character? Explain what elements or aspects of the character's design may have contributed to its popularity. \\
\midrule
LLaVA-v1.5-7B & The character in the image is a young boy with a distinctive appearance, wearing glasses, a bow tie, and a suit. This unique combination of elements, along with his smiling expression, suggests that the character is friendly, approachable, and confident. The character's design, which includes the bow tie and glasses, adds a touch of sophistication and charm to his overall look. These elements may have contributed to the character's popularity, as they make him stand out and create a memorable impression. The character's personality and design have likely resonated with audiences, making him a beloved and recognizable figure in popular culture. \\
\midrule
VARGPT & The character's personality seems to be confident, intelligent, and perhaps a bit mischievous, as suggested by the slight smirk and the way the eyes are directed slightly upwards and to the side, giving an impression of a playful or slightly rebellious attitude. The glasses and bow tie suggest a formal or professional setting, which could imply that the character is a well-educated individual. The design elements that may have contributed to its popularity include the expressive eyes, the confident stance, and the stylish yet formal attire.\\
\midrule
VARGPT-v1.1 & The character's personality appears to be that of a clever, observant detective, as suggested by the glasses and the serious expression. The large, expressive eyes and the confident smile could indicate a sense of intelligence and a quick wit. The character's popularity might be attributed to its distinctive design, which includes the glasses and bow tie, making it easily recognizable and endearing to fans of the series.\\
\bottomrule
\end{tabular}
}
\captionof{table}{The case of visual understanding in VARGPT-v1.1 shows that our VARGPT-v1.1 has achieved better understanding performance.}
\label{tab:tricky_example-3}  
  \end{minipage}
\end{table}

\begin{table}
  \begin{minipage}{0.99\linewidth}
\centering
\scalebox{0.80}{
\begin{tabular}{l p{8.0cm} }
\toprule
 \multicolumn{2}{l}{\bf Visual input example:}  \\
\midrule
&  \includegraphics[height=2.0cm]{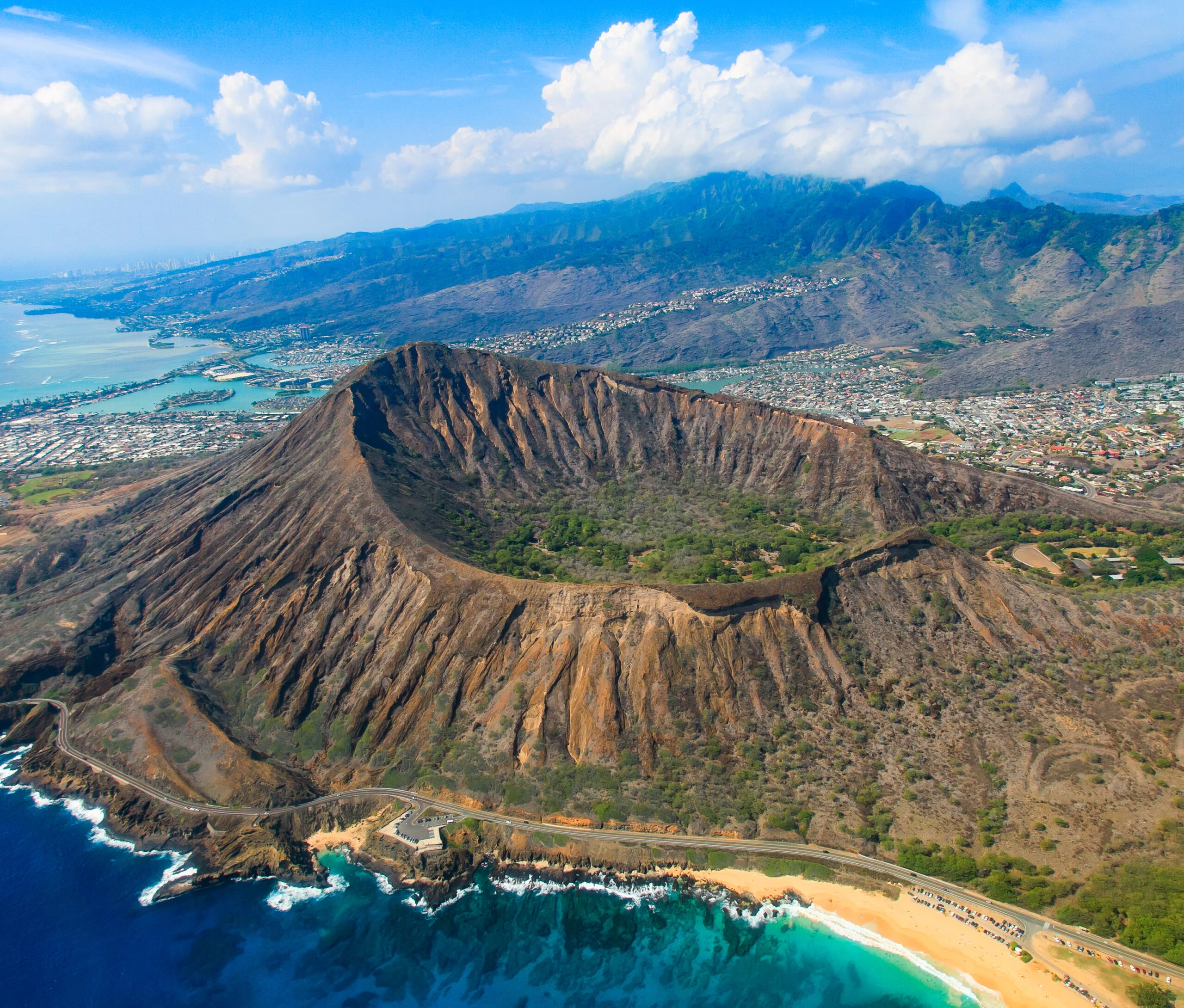} \\

User & What are the possible reasons of the formation of this sight? \\
\midrule
LLaVA-v1.5-7B & The formation of the sight, which is a large, rocky mountain with a lush green hillside and a beach, can be attributed to several factors. Firstly, the mountain's formation is likely due to tectonic forces, such as the collision or separation of tectonic plates, which have resulted in the formation of the mountain range. Secondly, the lush green hillside and the beach are the result of the local climate and geological processes. The presence of a beach suggests that the area is near a coastline, and the lush green hillside could be due to the presence of a water source, such as a river or a waterfall, which has created a fertile environment for plant growth. Additionally, the mountain's unique shape and the presence of a road on its side could be the result of human intervention, such as the construction of roads and infrastructure to facilitate transportation and access to the area. \\
\midrule
VARGPT & The sight of a large mountain with a road running through it and a beach nearby is a result of geological processes that have shaped the landscape over time. Some possible reasons for the formation of this sight include:1. Tectonic forces: The movement of the Earth's crust can lead to the formation of mountains, valleys, and other geological features. The mountain could have been formed through tectonic forces such as the collision of two tectonic plates or the uplift of the Earth's crust.2. Weathering and erosion: The mountain could have been shaped by the forces of weathering and erosion, which break down the rock and soil, leaving behind a rugged landscape. The road running through the mountain could be a result of human intervention to connect different areas or to facilitate transportation.3. Coastal erosion: The beach and the surrounding area could have been shaped by the forces of coastal erosion, which is the process of the sea wearing away the land. The beach could have been formed by the deposition of sediments from the sea, while the surrounding area could have been shaped by the erosion of rock and soil.4. Climate and weather patterns: The vegetation on the mountain and the surrounding area could be influenced by the local climate and weather patterns, which can affect the growth and distribution of plants and animals in the region.\\
\midrule
VARGPT-v1.1 & The sight in the image is a volcanic crater, which is formed by the eruption of a volcano. Volcanic eruptions can create craters of varying sizes and shapes, depending on the type of eruption and the materials involved. The crater in the image is likely the result of a past volcanic eruption that excavated a large cavity in the Earth's surface. Over time, the crater may have been shaped by erosion and weathering processes, and it may now be surrounded by vegetation and other geological features. The presence of the road and buildings around the crater suggests that it is a location of interest, possibly a tourist attraction or a natural landmark.\\
\bottomrule
\end{tabular}
}
\captionof{table}{The case of visual understanding in VARGPT-v1.1 shows that our VARGPT-v1.1 has achieved better understanding performance.}
\label{tab:tricky_example-1}  
  \end{minipage}
\end{table}

\begin{table}
  \begin{minipage}{0.99\linewidth}
\centering
\scalebox{0.80}{
\begin{tabular}{l p{7.5cm} }
\toprule
 \multicolumn{2}{l}{\bf Visual input example:}  \\
\midrule
&  \includegraphics[height=6cm]{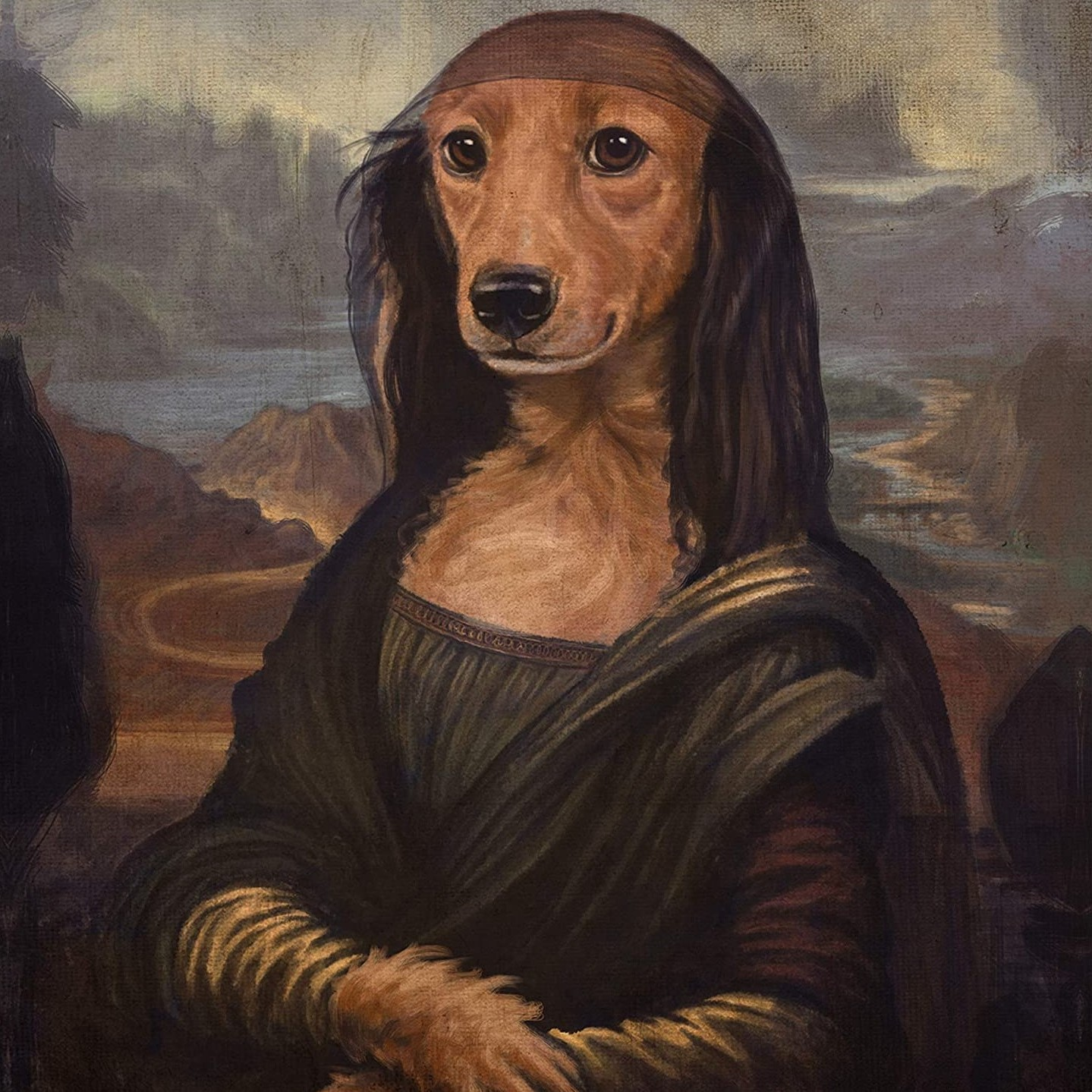} \\

User & Discuss how this creative twist on a classic work of art might be interpreted differently by various audiences. \\
\midrule
LLaVA-v1.5-7B & The creative twist on a classic work of art, featuring a dog wearing a Renaissance-style dress and a beret, can be interpreted differently by various audiences. Some might find it amusing and lighthearted, as it combines the seriousness of a historical painting with the playfulness of a dog dressed up in human attire. Others might view it as a commentary on the human-animal relationship, suggesting that dogs can be just as expressive and creative as humans. Additionally, some might perceive it as a critique of the traditional art world, as it challenges the norms of what is considered \"art\" by incorporating a non-human subject and a more contemporary, humorous element. Overall, the interpretation of this creative twist on a classic work of art can vary greatly depending on the viewer's perspective and personal experiences. \\
\midrule
VARGPT & This creative twist on a classic work of art, where a dog is dressed in a Mona Lisa-style outfit and painted in a similar style, could be interpreted in various ways by different audiences. Some might find it amusing and lighthearted, as it plays on the juxtaposition of a serious artwork with a playful, everyday subject. Others might appreciate the skillful use of artistic techniques to transform a simple image of a dog into a piece of art that can be appreciated on multiple levels. Some might even see it as a commentary on the universal appeal of art, which can be appreciated by both humans and animals. The interpretation would depend on the viewer's personal experiences, artistic background, and sense of humor.\\
\midrule
VARGPT-v1.1 & This creative twist on a classic work of art, featuring a dog with a human-like face, can be interpreted differently by various audiences. Some may find it humorous and entertaining, appreciating the cleverness of the artist in blending two distinct elements. Others might view it as a commentary on the nature of identity and the human-animal relationship, suggesting that even animals can possess human-like qualities or emotions. Some may also see it as a form of modern art, exploring the boundaries between different genres and challenging traditional notions of beauty and representation. Ultimately, the interpretation of this artwork will vary based on individual perspectives and cultural backgrounds.\\
\bottomrule
\end{tabular}
}
\captionof{table}{The case of visual understanding in VARGPT-v1.1 shows that our VARGPT-v1.1 has achieved better understanding performance.}
\label{tab:tricky_example-2}  
  \end{minipage}
\end{table}

\begin{table}
  \begin{minipage}{0.99\linewidth}
\centering
\scalebox{0.80}{
\begin{tabular}{l p{7.5cm} }
\toprule
 \multicolumn{2}{l}{\bf Visual input example:}  \\
\midrule
&  \includegraphics[height=4cm]{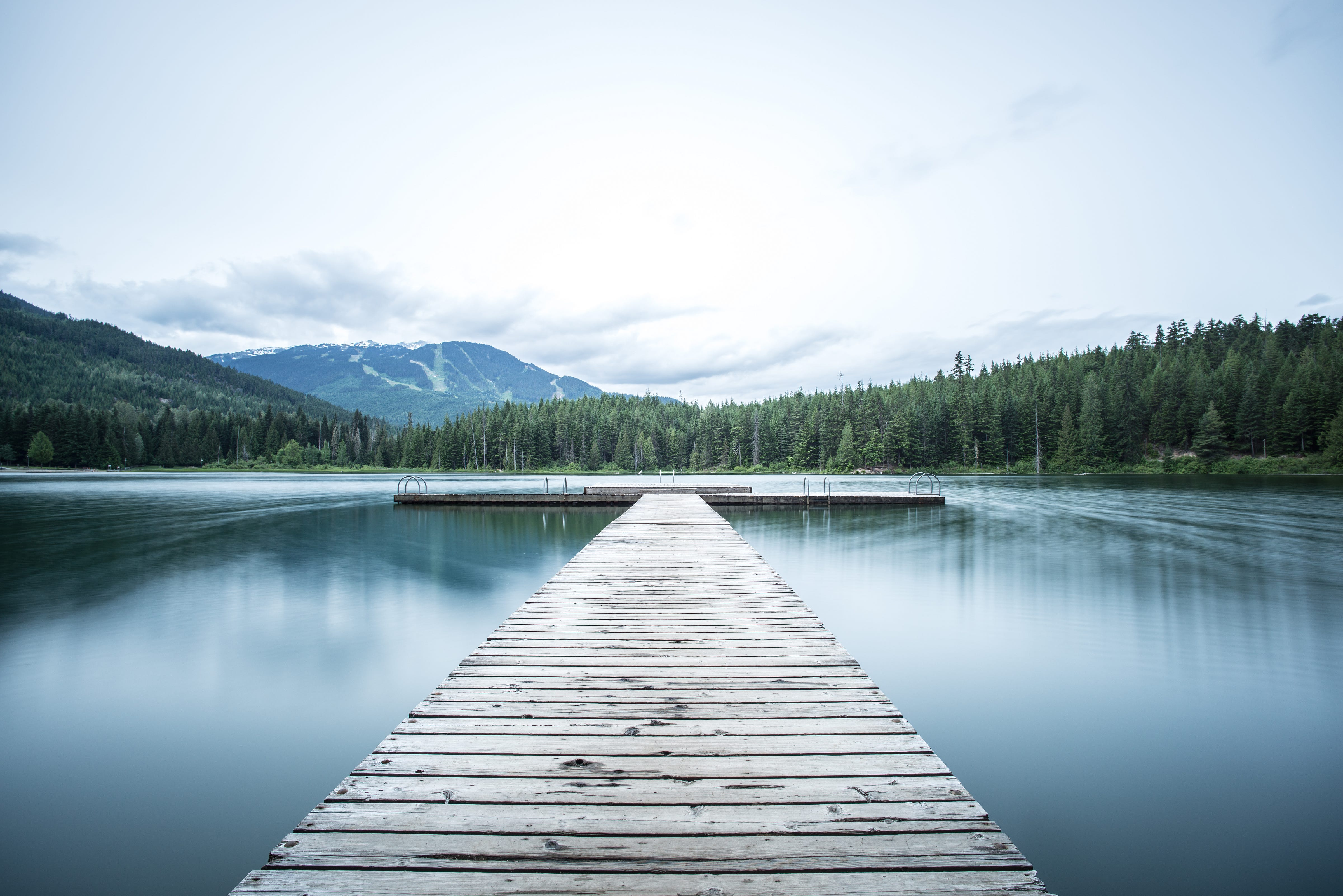} \\

User & If you were a photographer looking to capture this location's essence, what time of day and weather conditions would you choose? Describe the reasons behind your choice. \\
\midrule
LLaVA-v1.5-7B & To capture the essence of this location, which features a pier extending over a lake with mountains in the background, I would choose to photograph during the golden hour, which is the time just after sunrise or before sunset. This time of day provides soft, warm light that enhances the natural beauty of the scene, creating a serene and picturesque atmosphere. Additionally, the golden hour often results in softer shadows and more dramatic lighting, which can accentuate the pier's structure and the surrounding landscape.As for the weather conditions, I would prefer a clear day with a gentle breeze, as it would provide a pleasant and comfortable environment for both the photographer and the subjects. A clear day with good visibility would allow for better composition and capture the stunning mountain backdrop in its full glory. A gentle breeze would also add a touch of dynamism to the scene, making it more visually engaging and interesting. \\
\midrule
VARGPT & I would choose to capture this location's essence during the early morning, specifically during the golden hour, which is the period shortly after sunrise or before sunset. This time of day is known for its soft, warm light and long shadows, which can enhance the tranquility and beauty of the scene. The weather conditions would be clear and calm, with a gentle breeze to create a sense of movement in the water and to soften the reflections on the water's surface. The overcast sky would add a sense of depth and contrast to the scene, while the mist or fog would create a serene and ethereal atmosphere.\\
\midrule
VARGPT-v1.1 & I would choose to capture this location during the early morning or late afternoon, when the light is soft and the sky is clear, to create a serene and tranquil atmosphere. The calm water would reflect the light beautifully, enhancing the overall composition of the image. The weather should be clear with minimal cloud cover to ensure the best visibility of the distant mountains and the lush greenery. This would allow for a crisp and vibrant representation of the landscape, with the wooden dock and the lake's surface acting as natural elements that draw the viewer's eye towards the distant scenery.\\
\bottomrule
\end{tabular}
}
\captionof{table}{The case of visual understanding in VARGPT-v1.1 shows that our VARGPT-v1.1 has achieved better understanding performance.}
\label{tab:tricky_example-4}  
  \end{minipage}
\end{table}

\begin{table}
  \begin{minipage}{0.99\linewidth}
\centering
\scalebox{0.80}{
\begin{tabular}{l p{7.5cm} }
\toprule
 \multicolumn{2}{l}{\bf Visual input example:}  \\
\midrule
&  \includegraphics[height=5cm]{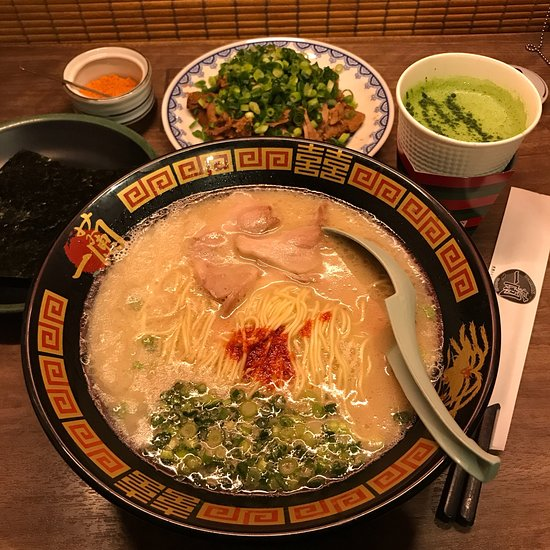} \\

User & If someone were to recommend a new flavor or topping to the dish, describe the reason for this change and how it might alter the overall taste. \\
\midrule
LLaVA-v1.5-7B & If someone were to recommend a new flavor or topping to the dish, they might suggest adding a spicy element, such as chili flakes or crushed red pepper, to enhance the taste and make it more flavorful. The current dish consists of a bowl of noodles with chicken, vegetables, and sauce, which is already a delicious and hearty meal. However, adding a spicy component could elevate the dish by providing a contrasting flavor and a kick of heat, making it more exciting and satisfying for those who enjoy spicy foods. This change would alter the overall taste by introducing a new dimension of flavor and complexity to the dish, making it more enjoyable and memorable for the person trying it. \\
\midrule
VARGPT & If someone were to recommend a new flavor or topping to the dish, it could be due to a desire to add a new taste dimension or to cater to personal preferences. For example, they might suggest adding a slice of avocado to the dish, which would introduce a creamy, buttery texture and a mild, nutty flavor that complements the richness of the noodles and the savory broth. This addition would alter the overall taste by providing a contrasting texture and a new layer of flavor complexity, enhancing the dish's overall appeal and making it more satisfying to eat.\\
\midrule
VARGPT-v1.1 & If someone were to recommend adding a new flavor or topping to the dish, it could be a type of meat, such as chicken or beef, to add more protein and depth to the broth. This would alter the overall taste by introducing a new flavor profile and texture, potentially making the dish more savory and satisfying.\\
\bottomrule
\end{tabular}
}
\captionof{table}{The case of visual understanding in VARGPT-v1.1 shows that our VARGPT-v1.1 has achieved better understanding performance.}
\label{tab:tricky_example-5}
  \end{minipage}
\end{table}

\newpage

{
    \small
    \bibliographystyle{ieeenat_fullname}
    \bibliography{main}
}


\end{document}